\documentclass[twoside,11pt]{article}
\usepackage{jair}
\usepackage{theapa}
\usepackage{amsmath}
\usepackage{multirow}
\makeatletter
\renewcommand{\jairheading}[5]{\def\ps@jairtps{\let\@mkboth\@gobbletwo%
\def\@oddhead{\scriptsize Journal of Artificial Intelligence Research #1 (#2) #3 \hfill Submitted #4; published #5}%
\def\@oddfoot{\scriptsize \copyright #2 National Research Council Canada.
Reprinted with permission. \hfill}%
\def\@evenhead{}\def\@evenfoot{}}%
\thispagestyle{jairtps}}
\makeatother
\jairheading{33}{2008}{615-655}{09/08}{12/08}
\ShortHeadings{The Latent Relation Mapping Engine}{Turney}
\firstpageno{615}
\makeatletter
\newcommand\figcaption{\def\@captype{figure}\caption}
\newcommand\tabcaption{\def\@captype{table}\caption}
\makeatother
\newenvironment{myitemize}{
\vspace{-0.3\baselineskip}
\begin{itemize}
  \setlength{\topsep}{0pt}
  \setlength{\itemsep}{3pt}
  \setlength{\parskip}{0pt}
  \setlength{\parsep}{0pt}
  \setlength{\partopsep}{0pt}
}{
\end{itemize}
\vspace{-0.2\baselineskip}}
\newenvironment{myenumerate}{
\vspace{-0.3\baselineskip}
\begin{enumerate}
  \setlength{\topsep}{0pt}
  \setlength{\itemsep}{3pt}
  \setlength{\parskip}{0pt}
  \setlength{\parsep}{0pt}
  \setlength{\partopsep}{0pt}
}{
\end{enumerate}
\vspace{-0.2\baselineskip}}
\begin{document}

\title{The Latent Relation Mapping Engine: \\
       Algorithm and Experiments}

\author{\name Peter D. Turney \email peter.turney@nrc-cnrc.gc.ca \\
       \addr Institute for Information Technology \\
       National Research Council Canada \\
       Ottawa, Ontario, Canada, K1A 0R6 }

\maketitle

\begin{abstract}
Many AI researchers and cognitive scientists have argued that analogy
is the core of cognition. The most influential work on computational
modeling of analogy-making is Structure Mapping Theory (SMT) and its
implementation in the Structure Mapping Engine (SME). A limitation
of SME is the requirement for complex hand-coded representations.
We introduce the Latent Relation Mapping Engine (LRME),
which combines ideas from SME and Latent Relational Analysis (LRA)
in order to remove the requirement for hand-coded representations.
LRME builds analogical mappings between lists of words, using
a large corpus of raw text to automatically discover the semantic
relations among the words. We evaluate LRME on a set of twenty
analogical mapping problems, ten based on scientific analogies and
ten based on common metaphors. LRME achieves human-level performance
on the twenty problems. We compare LRME with a variety of alternative
approaches and find that they are not able to reach the same
level of performance.
\end{abstract}

%
%

\section{Introduction}
\label{sec:intro}

When we are faced with a problem, we try to recall similar problems that
we have faced in the past, so that we can transfer our knowledge from
past experience to the current problem. We make an analogy between the
past situation and the current situation, and we use the analogy to transfer
knowledge \shortcite{gentner83,minsky86,holyoak95,hofstadter01,hawkins04}.

In his survey of the computational modeling of analogy-making, French
\citeyear{french02} cites Structure Mapping Theory (SMT) \shortcite{gentner83}
and its implementation in the Structure Mapping Engine (SME)
\cite{falkenhainer89} as the most influential work on modeling of
analogy-making. In SME, an analogical mapping $M: A \rightarrow B$ is from a source
$A$ to a target $B$. The source is more familiar, more known, or more concrete,
whereas the target is relatively unfamiliar, unknown, or abstract. The
analogical mapping is used to transfer knowledge from the source to the target.

Gentner \citeyear{gentner83} argues that there are two kinds of similarity,
attributional similarity and relational similarity. The distinction between
attributes and relations may be understood in terms of predicate logic. An
attribute is a predicate with one argument, such as {\sc large}($X$), meaning
$X$ is large. A relation is a predicate with two or more arguments, such
as {\sc collides\_with}($X,Y$), meaning $X$ collides with $Y$.

The Structure Mapping Engine prefers mappings based on relational
similarity over mappings based on attributional similarity \shortcite{falkenhainer89}.
For example, SME is able to build a mapping from a representation of the
solar system (the source) to a representation of the Rutherford-Bohr model
of the atom (the target). The sun is mapped to the nucleus, planets are
mapped to electrons, and mass is mapped to charge.
Note that this mapping emphasizes relational similarity. The sun and the
nucleus are very different in terms of their attributes: the sun is
very large and the nucleus is very small. Likewise, planets and electrons
have little attributional similarity. On the other hand, planets revolve
around the sun like electrons revolve around the nucleus. The mass of the
sun attracts the mass of the planets like the charge of the nucleus attracts
the charge of the electrons.

Gentner \citeyear{gentner91} provides evidence that children rely primarily
on attributional similarity for mapping, gradually switching over to
relational similarity as they mature. She uses the terms
{\em mere appearance} to refer to mapping based mostly on
attributional similarity, {\em analogy} to refer to mapping based
mostly on relational similarity, and {\em literal similarity} to
refer to a mixture of attributional and relational similarity.
Since we use analogical mappings to solve problems and make
predictions, we should focus on structure, especially causal relations,
and look beyond the surface attributes of things \shortcite{gentner83}.
The analogy between the solar system and the Rutherford-Bohr model
of the atom illustrates the importance of going beyond mere appearance, to
the underlying structures.

Figures \ref{fig:solar} and \ref{fig:atom} show the LISP representations
used by SME as input for the analogy between the solar system and the
atom \shortcite{falkenhainer89}. Chalmers, French, and Hofstadter
\citeyear{chalmers92} criticize SME's requirement for complex hand-coded
representations. They argue that most of the hard work is done by the human who
creates these high-level hand-coded representations, rather than by SME.

\begin{table}[htbp]
\footnotesize
\tt
\centering
\begin{tabular}{|l|}
\hline
(defEntity sun :type inanimate) \\
(defEntity planet :type inanimate) \\
\\
(defDescription solar-system \\
\hspace{20pt} entities (sun planet) \\
\hspace{20pt} expressions  (((mass sun) :name mass-sun) \\
\hspace{40pt} ((mass planet) :name mass-planet) \\
\hspace{40pt} ((greater mass-sun mass-planet) :name >mass) \\
\hspace{40pt} ((attracts sun planet) :name attracts-form) \\
\hspace{40pt} ((revolve-around planet sun) :name revolve) \\
\hspace{40pt} ((and >mass attracts-form) :name and1) \\
\hspace{40pt} ((cause and1 revolve) :name cause-revolve) \\
\hspace{40pt} ((temperature sun) :name temp-sun) \\
\hspace{40pt} ((temperature planet) :name temp-planet) \\
\hspace{40pt} ((greater temp-sun temp-planet) :name >temp) \\
\hspace{40pt} ((gravity mass-sun mass-planet) :name force-gravity) \\
\hspace{40pt} ((cause force-gravity attracts-form) :name why-attracts))) \\
\hline
\end{tabular}
\normalsize
\rm
\figcaption {The representation of the solar system in SME
\shortcite{falkenhainer89}.}
\label{fig:solar}
\end{table}

\begin{table}[htbp]
\footnotesize
\tt
\centering
\begin{tabular}{|l|}
\hline
(defEntity nucleus :type inanimate) \\
(defEntity electron :type inanimate) \\
\\
(defDescription rutherford-atom \\
\hspace{20pt} entities (nucleus electron) \\
\hspace{20pt} expressions  (((mass nucleus) :name mass-n) \\
\hspace{40pt} ((mass electron) :name mass-e) \\
\hspace{40pt} ((greater mass-n mass-e) :name >mass) \\
\hspace{40pt} ((attracts nucleus electron) :name attracts-form) \\
\hspace{40pt} ((revolve-around electron nucleus) :name revolve) \\
\hspace{40pt} ((charge electron) :name q-electron) \\
\hspace{40pt} ((charge nucleus) :name q-nucleus) \\
\hspace{40pt} ((opposite-sign q-nucleus q-electron) :name >charge) \\
\hspace{40pt} ((cause >charge attracts-form) :name why-attracts))) \\
\hline
\end{tabular}
\normalsize
\rm
\figcaption {The Rutherford-Bohr model of the atom
in SME \shortcite{falkenhainer89}.}
\label{fig:atom}
\end{table}

Gentner, Forbus, and their colleagues have attempted to avoid hand-coding
in their recent work with SME.\footnote{Dedre Gentner, personal communication,
October 29, 2008.} The CogSketch system can generate LISP representations
from simple sketches \cite{forbus08}. The Gizmo system can generate
LISP representations from qualitative physics models \cite{yan05}. The
Learning Reader system can generate LISP representations from natural language
text \shortcite{forbus07}. These systems do not require LISP input.

However, the CogSketch user interface requires the person who draws the
sketch to identify the basic components in the sketch and hand-label
them with terms from a knowledge base derived from OpenCyc.
Forbus et al. \citeyear{forbus08} note that OpenCyc
contains more than 58,000 hand-coded concepts, and they have added further
hand-coded concepts to OpenCyc, in order to support CogSketch.
The Gizmo system requires the user to hand-code a physical model,
using the methods of qualitative physics \shortcite{yan05}. Learning Reader
uses more than 28,000 phrasal patterns, which were derived from
ResearchCyc \shortcite{forbus07}. It is evident that SME still requires
substantial hand-coded knowledge.

The work we present in this paper is an effort to avoid
complex hand-coded representations.
Our approach is to combine ideas from SME \shortcite{falkenhainer89}
and Latent Relational Analysis (LRA) \shortcite{turney06}. We call
the resulting algorithm the Latent Relation Mapping Engine (LRME).
We represent the semantic relation between two terms using a
vector, in which the elements are derived from
pattern frequencies in a large corpus of raw text. Because the
semantic relations are automatically derived from a corpus, LRME
does not require hand-coded representations of relations. It only
needs a list of terms from the source and a list of terms from
the target. Given these two lists, LRME uses the corpus to build
representations of the relations among the terms, and then it
constructs a mapping between the two lists.

Tables \ref{tab:input} and
\ref{tab:output} show the input and output of LRME for
the analogy between the solar system and the Ruther\-ford-Bohr model of
the atom. Although some human effort is involved in constructing
the input lists, it is considerably less effort than SME requires for its input
(contrast Figures \ref{fig:solar} and \ref{fig:atom} with Table~\ref{tab:input}).

\begin{table}[htbp]
\centering
\begin{minipage}{0.3\textwidth}
\centering
\begin{tabular}{l}
\hline
\textbf{Source $A$} \\
\hline
planet \\
attracts \\
revolves \\
sun \\
gravity \\
solar system  \\
mass  \\
\hline
\end{tabular}
\end{minipage}
\begin{minipage}{0.3\textwidth}
\centering
\begin{tabular}{l}
\hline
\textbf{Target $B$} \\
\hline
revolves \\
atom \\
attracts \\
electromagnetism \\
nucleus \\
charge \\
electron \\
\hline
\end{tabular}
\end{minipage}
\caption {The representation of the input in LRME.}
\label{tab:input}
\end{table}

\begin{table}[htbp]
\centering
\begin{tabular}{lcl}
\hline
\textbf{Source $A$} & \textbf{Mapping $M$} & \textbf{Target $B$} \\
\hline
solar system       & $\rightarrow$  & atom \\
sun                & $\rightarrow$  & nucleus \\
planet             & $\rightarrow$  & electron \\
mass               & $\rightarrow$  & charge \\
attracts           & $\rightarrow$  & attracts \\
revolves           & $\rightarrow$  & revolves \\
gravity            & $\rightarrow$  & electromagnetism \\
\hline
\end{tabular}
\caption {The representation of the output in LRME.}
\label{tab:output}
\end{table}

Scientific analogies, such as the analogy between the solar system and
the Rutherford-Bohr model of the atom, may seem esoteric, but we believe
analogy-making is ubiquitous in our daily lives. A potential practical
application for this work is the task of identifying semantic roles
\shortcite{gildea02}. Since roles are relations, not attributes, it is
appropriate to treat semantic role labeling as an analogical mapping problem.

For example, the {\sc Judgement} semantic frame contains
semantic roles such as {\sc judge}, {\sc evaluee}, and {\sc reason}, and the
{\sc Statement} frame contains roles such as {\sc speaker}, {\sc addressee},
{\sc message}, {\sc topic}, and {\sc medium} \shortcite{gildea02}. The task
of identifying semantic roles is to automatically label sentences with their
roles, as in the following examples \shortcite{gildea02}:

\begin{myitemize}

\item $[${\em Judge} She] \textbf{blames} [{\em Evaluee} the Government]
[{\em Reason} for failing to do enough to help].

\item $[${\em Speaker} We] \textbf{talked} [{\em Topic} about the proposal]
[{\em Medium} over the phone].

\end{myitemize}

\noindent If we have a training set of labeled sentences and a testing
set of unlabeled sentences, then we may view the task of labeling the
testing sentences as a problem of creating analogical mappings between
the training sentences (sources) and the testing sentences (targets).
Table~\ref{tab:roles} shows how ``She blames the Government
for failing to do enough to help.'' might be mapped to ``They blame the
company for polluting the environment.'' Once a mapping has been found,
we can transfer knowledge, in the form of semantic role labels, from
the source to the target.

\begin{table}[htbp]
\centering
\begin{tabular}{lcl}
\hline
\textbf{Source $A$} & \textbf{Mapping $M$} & \textbf{Target $B$} \\
\hline
she                & $\rightarrow$  & they \\
blames             & $\rightarrow$  & blame \\
government         & $\rightarrow$  & company \\
failing            & $\rightarrow$  & polluting \\
help               & $\rightarrow$  & environment \\
\hline
\end{tabular}
\caption {Semantic role labeling as analogical mapping.}
\label{tab:roles}
\end{table}

In Section~\ref{sec:hypotheses}, we briefly discuss the hypotheses behind
the design of LRME. We then precisely define the task that is performed
by LRME, a specific form of analogical mapping, in Section~\ref{sec:task}.
LRME builds on Latent Relational Analysis (LRA), hence we summarize
LRA in Section~\ref{sec:lra}. We discuss potential applications of LRME in
Section~\ref{sec:apps}.

To evaluate LRME, we created twenty analogical mapping problems, ten
science analogy problems \shortcite{holyoak95} and ten common metaphor problems
\shortcite{lakoff80}. Table~\ref{tab:input} is one of the science analogy
problems. Our intended solution is given in Table~\ref{tab:output}.
To validate our intended solutions, we gave our colleagues the lists of terms
(as in Table~\ref{tab:input}) and asked them to generate mappings between the lists.
Section~\ref{sec:problems} presents the results of this experiment.
Across the twenty problems, the average agreement with our intended
solutions (as in Table~\ref{tab:output}) was 87.6\%.

The LRME algorithm is outlined in Section~\ref{sec:lrme}, along with its
evaluation on the twenty mapping problems. LRME achieves an accuracy
of 91.5\%. The difference between this performance and the human average
of 87.6\% is not statistically significant.

Section~\ref{sec:attributes} examines a variety of alternative approaches
to the analogy mapping task. The best approach achieves an accuracy of
76.8\%, but this approach requires hand-coded part-of-speech tags.
This performance is significantly below LRME and human performance.

In Section~\ref{sec:discussion}, we discuss some questions that are
raised by the results in the preceding sections. Related work is
described in Section~\ref{sec:related}, future work and limitations
are considered in Section~\ref{sec:future}, and we conclude in
Section~\ref{sec:conclusion}.

%
%

\section{Guiding Hypotheses}
\label{sec:hypotheses}

In this section, we list some of the assumptions that have guided the
design of LRME. The results we present in this paper do not necessarily
require these assumptions, but it might be helpful to the reader, to
understand the reasoning behind our approach.

\begin{myenumerate}

\item \textbf{Analogies and semantic relations:} Analogies are
based on semantic relations \shortcite{gentner83}. For example, the analogy
between the solar system and the Ruther\-ford-Bohr model of the atom is
based on the similarity of the semantic relations among the concepts
involved in our understanding of the solar system to the semantic
relations among the concepts involved in the Ruther\-ford-Bohr model of the atom.

\item \textbf{Co-occurrences and semantic relations:} Two terms have an
interesting, significant semantic relation if and only if they they tend
to co-occur within a relatively small window (e.g., five words) in a relatively
large corpus (e.g., $10^{10}$ words). Having an interesting semantic relation causes
co-occurrence and co-occurrence is a reliable indicator of an interesting
semantic relation \shortcite{firth57}.

\item \textbf{Meanings and semantic relations:} Meaning has more to do with
relations among words than individual words. Individual words tend
to be ambiguous and polysemous. By putting two words into a pair, we constrain
their possible meanings. By putting words into a sentence, with multiple
relations among the words in the sentence, we constrain the possible meanings
further. If we focus on word pairs (or tuples), instead of individual words,
word sense disambiguation is less problematic. Perhaps a word has no sense
apart from its relations with other words \shortcite{kilgarriff97}.

\item \textbf{Pattern distributions and semantic relations:} There is a
many-to-many mapping between semantic relations and the patterns in which
two terms co-occur. For example, the relation ${\rm CauseEffect}(X,Y)$ may be
expressed as ``$X$ causes $Y$'', ``$Y$ from $X$'', ``$Y$ due to $X$'',
``$Y$ because of $X$'', and so on. Likewise, the pattern ``$Y$ from $X$''
may be an expression of ${\rm CauseEffect}(X,Y)$ (``sick from bacteria'')
or ${\rm OriginEntity}(X,Y)$ (``oranges from Spain''). However,
for a given $X$ and $Y$, the statistical distribution of patterns in which $X$ and $Y$
co-occur is a reliable signature of the semantic relations between $X$ and $Y$
\shortcite{turney06}.

\end{myenumerate}

\noindent To the extent that LRME works, we believe its success lends some
support to these hypotheses.

%
%

\section{The Task}
\label{sec:task}

In this paper, we examine algorithms that generate analogical mappings.
For simplicity, we restrict the task to generating {\em bijective} mappings;
that is, mappings that are both {\em injective} (one-to-one; there is no
instance in which two terms in the source map to the same term in the target)
and {\em surjective} (onto; the source terms cover all of the target terms;
there is no target term that is left out of the mapping). We assume that
the entities that are to be mapped are given as input. Formally, the input
$I$ for the algorithms is two sets of terms, $A$ and $B$.

\begin{equation}
I = \left \{ \left \langle A, B \right \rangle \right \}
\end{equation}

\noindent Since the mappings are bijective, $A$ and $B$ must contain
the same number of terms, $m$.

\begin{align}
A & = \left \{ a_1, a_2, \ldots, a_m \right \} \\
B & = \left \{ b_1, b_2, \ldots, b_m \right \}
\end{align}

\noindent A term, $a_i$ or $b_j$, may consist of a single word ({\em planet}) or
a compound of two or more words ({\em solar system}). The words may be any
part of speech (nouns, verbs, adjectives, or adverbs). The output $O$ is a
bijective mapping $M$ from $A$ to $B$.

\begin{align}
O & = \left \{ M: A \rightarrow B \right \} \\
M(a_i) & \in B \\
M(A) & = \left \{ M(a_1), M(a_2), \ldots, M(a_m) \right \} = B
\end{align}

\noindent The algorithms that we consider here
can accept a batch of multiple independent mapping problems
as input and generate a mapping for each one as output.

\begin{align}
I & = \left \{ \left \langle A_1, B_1 \right \rangle,
\left \langle A_2, B_2 \right \rangle, \ldots,
\left \langle A_n, B_n \right \rangle \right \} \\
O & = \left \{ M_1: A_1 \rightarrow B_1,
M_2: A_2 \rightarrow B_2, \ldots,
M_n: A_n \rightarrow B_n \right \}
\end{align}

Suppose the terms in $A$ are in some arbitrary order $\mathbf{a}$.

\begin{equation}
\mathbf{a} = \left \langle a_1, a_2, \ldots, a_m \right \rangle
\end{equation}

\noindent The mapping function $M: A \rightarrow B$, given $\mathbf{a}$,
determines a unique ordering $\mathbf{b}$ of $B$.

\begin{equation}
\mathbf{b} = \left \langle M(a_1), M(a_2), \ldots, M(a_m) \right \rangle
\end{equation}

\noindent Likewise, an ordering $\mathbf{b}$ of $B$, given $\mathbf{a}$,
defines a unique mapping function $M$. Since there are $m!$ possible
orderings of $B$, there are also $m!$ possible mappings from $A$ to $B$.
The task is to search through the $m!$ mappings and find the best one.
(Section~\ref{sec:problems} shows that there is a relatively high
degree of consensus about which mappings are best.)

Let $P(A,B)$ be the set of all $m!$ bijective mappings from $A$ to $B$.
($P$ stands for {\em permutation}, since each mapping corresponds to
a permutation.)

\begin{align}
P(A,B) & = \left \{ M_1, M_2, \ldots, M_{m!} \right \} \\
m & = \left | A \right | = \left | B \right | \\
m! & = \left | P(A,B) \right |
\end{align}

\noindent In the following experiments, $m$ is $7$ on average and $9$
at most, so $m!$ is usually around $7! = 5,040$ and at most $9! = 362,880$.
It is feasible for us to exhaustively search $P(A,B)$.

We explore two basic kinds of algorithms for generating analogical
mappings, algorithms based on {\em attributional similarity} and
algorithms based on {\em relational similarity} \shortcite{turney06}.
The attributional similarity between two words, ${\rm sim_a}(a,b) \in \Re$,
depends on the degree of correspondence between the properties of $a$ and $b$.
The more correspondence there is, the greater their attributional similarity.
The relational similarity between two {\em pairs} of words,
${\rm sim_r}(a\!:\!b, c\!:\!d) \in \Re$, depends on the degree of correspondence
between the relations of $a\!:\!b$ and $c\!:\!d$. The more correspondence there is,
the greater their relational similarity. For example, {\em dog} and {\em wolf}
have a relatively high degree of attributional similarity, whereas
{\em dog}$\,:\,${\em bark} and {\em cat}$\,:\,${\em meow} have a relatively
high degree of relational similarity.

Attributional mapping algorithms seek the mapping (or mappings) $M_{\rm a}$ that
maximizes the sum of the attributional similarities between the terms in $A$
and the corresponding terms in $B$. (When there are multiple mappings that
maximize the sum, we break the tie by randomly choosing one of them.)

\begin{equation}
\label{eqn:att-alg}
M_{\rm a} = \operatornamewithlimits{arg\,max}_{M \in P(A,B)}
\; \sum_{i=1}^{m} {\rm sim_a}(a_i, M(a_i))
\end{equation}

Relational mapping algorithms seek the mapping (or mappings) $M_{\rm r}$ that
maximizes the sum of the relational similarities.

\begin{equation}
\label{eqn:rel-alg}
M_{\rm r} = \operatornamewithlimits{arg\,max}_{M \in P(A,B)}
\; \sum_{i=1}^{m} \sum_{j=i+1}^{m} {\rm sim_r}(a_i\!:\!a_j, M(a_i)\!:\!M(a_j))
\end{equation}

\noindent In (\ref{eqn:rel-alg}), we assume that ${\rm sim_r}$ is
symmetrical. For example, the degree of relational similarity between
{\em dog}$\,:\,${\em bark} and {\em cat}$\,:\,${\em meow} is the
same as the degree of relational similarity between
{\em bark}$\,:\,${\em dog} and {\em meow}$\,:\,${\em cat}.

\begin{equation}
\label{eqn:symmetry}
{\rm sim_r}(a\!:\!b, c\!:\!d) = {\rm sim_r}(b\!:\!a, d\!:\!c)
\end{equation}

\noindent We also assume that ${\rm sim_r}(a\!:\!a, b\!:\!b)$ is
not interesting; for example, it may be some constant value for
all $a$ and $b$. Therefore (\ref{eqn:rel-alg}) is designed so that
$i$ is always less than $j$.

Let ${\rm score_r}(M)$ and ${\rm score_a}(M)$ be defined as follows.

\begin{align}
\label{eqn:score-r}
{\rm score_r}(M)
& = \sum_{i=1}^{m} \sum_{j=i+1}^{m} {\rm sim_r}(a_i\!:\!a_j, M(a_i)\!:\!M(a_j)) \\
\label{eqn:score-a}
{\rm score_a}(M)
& = \sum_{i=1}^{m} {\rm sim_a}(a_i, M(a_i))
\end{align}

\noindent Now $M_{\rm r}$ and $M_{\rm a}$ may be defined in terms
of ${\rm score_r}(M)$ and ${\rm score_a}(M)$.

\begin{align}
M_{\rm r}
& = \operatornamewithlimits{arg\,max}_{M \in P(A,B)} {\rm score_r}(M) \\
M_{\rm a}
& = \operatornamewithlimits{arg\,max}_{M \in P(A,B)} {\rm score_a}(M)
\end{align}

\noindent $M_{\rm r}$ is the best mapping according to
${\rm sim_r}$ and $M_{\rm a}$ is the best mapping according to
${\rm sim_a}$.

Recall Gentner's \citeyear{gentner91} terms, discussed in
Section~\ref{sec:intro}, {\em mere appearance} (mostly attributional
similarity), {\em analogy} (mostly relational similarity), and
{\em literal similarity} (a mixture of attributional and relational
similarity). We take it that $M_{\rm r}$ is an abstract model of
mapping based on analogy and $M_{\rm a}$ is a model
of mere appearance. For literal similarity, we can combine
$M_{\rm r}$ and $M_{\rm a}$, but we should take care to normalize
${\rm score_r}(M)$ and ${\rm score_a}(M)$ before we combine them.
(We experiment with combining them in Section~\ref{subsec:hybrids}.)

%
%

\section{Latent Relational Analysis}
\label{sec:lra}

LRME uses a simplified form of Latent Relational Analysis (LRA)
\shortcite{turney05b,turney06} to calculate the relational similarity
between pairs of words. We will briefly describe past work with LRA before
we present LRME.

LRA takes as input $I$ a set of word pairs and generates as output $O$ the
relational similarity ${\rm sim_r}(a_i\!:\!b_i, a_j\!:\!b_j)$ between any
two pairs in the input.

\begin{align}
I & = \left \{  a_1\!:\!b_1, a_2\!:\!b_2, \ldots, a_n\!:\!b_n \right \} \\
O & = \left \{ {\rm sim_r} : I \times I \rightarrow \Re \right \}
\end{align}

\noindent LRA was designed to evaluate proportional analogies.
Proportional analogies have the form $a\!:\!b\!::\!c\!:\!d$,
which means ``$a$ is to $b$ as $c$ is to $d$''. For example,
{\em mason}$\,:\,${\em stone}$\,::\,${\em carpenter}$\,:\,${\em wood}
means ``mason is to stone as carpenter is to wood''. A mason is an artisan
who works with stone and a carpenter is an artisan who works with wood.

We consider proportional analogies to be a special case of bijective
analogical mapping, as defined in Section~\ref{sec:task}, in which
$\left | A \right | = \left | B \right | = m = 2$.
For example, $a_1\!:\!a_2\!::\!b_1\!:\!b_2$ is equivalent to $M_0$ in
(\ref{eqn:ab2}).

\begin{equation}
\label{eqn:ab2}
A = \left \{ a_1, a_2 \right \},\;
B = \left \{ b_1, b_2 \right \},\;
M_0(a_1) = b_1,\;
M_0(a_2) = b_2.
\end{equation}

\noindent From the definition of ${\rm score_r}(M)$ in
(\ref{eqn:score-r}), we have the following result for $M_0$.

\begin{equation}
\label{eqn:rel-qual}
{\rm score_r}(M_0) =
{\rm sim_r}(a_1\!:\!a_2, M_0(a_1)\!:\!M_0(a_2)) =
{\rm sim_r}(a_1\!:\!a_2, b_1\!:\!b_2)
\end{equation}

\noindent That is, the quality of the proportional analogy
{\em mason}$\,:\,${\em stone}$\,::\,${\em carpenter}$\,:\,${\em wood}
is given by ${\rm sim_r}(mason\!:\!stone, carpenter\!:\!wood)$.

Proportional analogies may also be evaluated using attributional
similarity. From the definition of ${\rm score_a}(M)$ in
(\ref{eqn:score-a}), we have the following result for $M_0$.

\begin{equation}
\label{eqn:att-qual}
{\rm score_a}(M_0) =
{\rm sim_a}(a_1, M_0(a_1)) + {\rm sim_a}(a_2, M_0(a_2)) =
{\rm sim_a}(a_1, b_1) + {\rm sim_a}(a_2, b_2)
\end{equation}

\noindent For attributional similarity, the quality of the
proportional analogy
{\em mason}$\,:\,${\em stone}$\,::\,${\em carpenter}$\,:\,${\em wood}
is given by ${\rm sim_a}(mason,carpenter) + {\rm sim_a}(stone,wood)$.

LRA only handles proportional analogies. The main contribution of LRME
is to extend LRA beyond proportional analogies to bijective analogies
for which $m > 2$.

Turney \citeyear{turney06} describes ten potential applications of
LRA: recognizing proportional analogies, structure
mapping theory, modeling metaphor, classifying semantic relations,
word sense disambiguation, information extraction, question answering,
automatic thesaurus generation, information retrieval, and
identifying semantic roles. Two of these applications (evaluating
proportional analogies and classifying semantic relations) are
experimentally evaluated, with state-of-the-art results.

Turney \citeyear{turney06} compares the performance of
relational similarity (\ref{eqn:rel-qual}) and attributional
similarity (\ref{eqn:att-qual}) on the task of solving 374
multiple-choice proportional analogy questions from the SAT college
entrance test. LRA is used to measure relational similarity and a
variety of lexicon-based and corpus-based algorithms are
used to measure attributional similarity. LRA
achieves an accuracy of 56\% on the 374 SAT questions, which
is not significantly different from the average human score
of 57\%. On the other hand, the best performance by attributional
similarity is 35\%. The results show that
attributional similarity is better than random guessing, but not
as good as relational similarity. This
result is consistent with Gentner's \citeyear{gentner91}
theory of the maturation of human similarity judgments.

Turney \citeyear{turney06} also applies LRA to the task of
classifying semantic relations in noun-modifier expressions.
A noun-modifier expression is a phrase, such as {\em laser
printer}, in which the head noun ({\em printer}) is preceded
by a modifier ({\em laser}). The task is to identify the semantic
relation between the noun and the modifier. In this case, the
relation is {\em instrument}; the laser is an {\em instrument}
used by the printer. On a set of 600 hand-labeled noun-modifier
pairs with five different classes of semantic relations, LRA
attains 58\% accuracy.

Turney \citeyear{turney08} employs a variation of LRA for solving
four different language tests, achieving 52\% accuracy on SAT
analogy questions, 76\% accuracy on TOEFL synonym questions,
75\% accuracy on the task of distinguishing synonyms from antonyms,
and 77\% accuracy on the task of distinguishing words that are
similar, words that are associated, and words that are both
similar and associated. The same core algorithm is used for
all four tests, with no tuning of the parameters to the particular
test.

%
%

\section{Applications for LRME}
\label{sec:apps}

Since LRME is an extension of LRA, every potential application of LRA is
also a potential application of LRME. The advantage of LRME over LRA is
the ability to handle bijective analogies when $m > 2$ (where
$m = \left | A \right | = \left | B \right |$). In this section,
we consider the kinds of applications that might benefit from this ability.

In Section~\ref{subsec:experiments}, we evaluate LRME on science analogies
and common metaphors, which supports the claim that these two applications
benefit from the ability to handle larger sets of terms. In
Section~\ref{sec:intro}, we saw that identifying semantic
roles \shortcite{gildea02} also involves more than two terms, and
we believe that LRME will be superior to LRA for semantic role labeling.

Semantic relation classification usually assumes that the relations
are binary; that is, a semantic relation is a connection between
two terms \shortcite{rosario01,nastase03,turney06,girju07}. Yuret
observed that binary relations may be linked by
underlying $n$-ary relations.\footnote{Deniz Yuret, personal
communication, February 13, 2007. This observation was in the context
of our work on building the datasets for SemEval 2007
Task 4 \shortcite{girju07}.} For example, Nastase and
Szpakowicz \citeyear{nastase03} defined a taxonomy of 30 binary semantic relations.
Table~\ref{tab:n-ary} shows how six binary relations from Nastase and
Szpakowicz \citeyear{nastase03} can be covered by one \mbox{5-ary} relation,
Agent:Tool:Action:Affected:Theme. An Agent uses a Tool to perform an Action.
Somebody or something is Affected by the Action. The whole event can be
summarized by its Theme.

\begin{table}[htbp]
\small
\centering
\begin{tabular}{lll}
\hline
\multicolumn{2}{c}{\textbf{Nastase and Szpakowicz \citeyear{nastase03}}} & \\
\cline{1-2}
\textbf{Relation} & \textbf{Example} & \textbf{Agent:Tool:Action:Affected:Theme} \\
\hline
agent           & student protest   & Agent:Action \\
purpose         & concert hall      & Theme:Tool \\
beneficiary     & student discount  & Affected:Action \\
instrument      & laser printer     & Tool:Agent \\
object          & metal separator   & Affected:Tool \\
object property & sunken ship       & Action:Affected \\
\hline
\end{tabular}
\normalsize
\caption {How six binary semantic relations from Nastase and
Szpakowicz \citeyear{nastase03} can be viewed as different fragments
of one \mbox{5-ary} semantic relation.}
\label{tab:n-ary}
\end{table}

In SemEval Task 4, we found it easier to manually tag the datasets
when we expanded binary relations to their underlying \mbox{$n$-ary}
relations \shortcite{girju07}. We believe that this expansion would
also facilitate automatic classification of semantic relations.
The results in Section~\ref{subsec:coherence} suggest that all of
the applications for LRA that we discussed in Section~\ref{sec:lra}
might benefit from being able to handle bijective analogies when $m > 2$.

%
%

\section{The Mapping Problems}
\label{sec:problems}

To evaluate our algorithms for analogical mapping, we created twenty mapping
problems, given in Appendix A. The twenty problems consist of ten science
analogy problems, based on examples of analogy in science from Chapter~8 of
Holyoak and Thagard \citeyear{holyoak95}, and ten common metaphor problems,
derived from Lakoff and Johnson \citeyear{lakoff80}.

The tables in Appendix A show our intended mappings for each of the twenty
problems. To validate these mappings, we invited our colleagues in the
Institute for Information Technology to participate in
an experiment. The experiment was hosted on a web server (only accessible
inside our institute) and people participated anonymously, using their web
browsers in their offices. There were 39 volunteers who began the
experiment and 22 who went all the way to the end. In our analysis,
we use only the data from the 22 participants who completed all of the
mapping problems.

The instructions for the participants are in Appendix A. The sequence
of the problems and the order of the terms within a problem were
randomized separately for each participant, to remove any effects
due to order. Table~\ref{tab:agreement} shows the agreement between
our intended mapping and the mappings generated by the participants.
Across the twenty problems, the average agreement was 87.6\%, which is
higher than the agreement figures for many linguistic annotation tasks.
This agreement is impressive, given that the participants had minimal
instructions and no training.

\begin{table}[htbp]
\small
\centering
\begin{tabular}{lllrl}
\hline
\textbf{Type} & \textbf{Mapping} & \textbf{Source $\rightarrow$ Target}
& \textbf{Agreement} & \textbf{$m$} \\
\hline
& A1  & solar system              $\rightarrow$ atom                      &  90.9 & 7 \\
& A2  & water flow                $\rightarrow$ heat transfer             &  86.9 & 8 \\
& A3  & waves                     $\rightarrow$ sounds                    &  81.8 & 8 \\
& A4  & combustion                $\rightarrow$ respiration               &  79.0 & 8 \\
science
& A5  & sound                     $\rightarrow$ light                     &  79.2 & 7 \\
analogies
& A6  & projectile                $\rightarrow$ planet                    &  97.4 & 7 \\
& A7  & artificial selection      $\rightarrow$ natural selection         &  74.7 & 7 \\
& A8  & billiard balls            $\rightarrow$ gas molecules             &  88.1 & 8 \\
& A9  & computer                  $\rightarrow$ mind                      &  84.3 & 9 \\
& A10 & slot machine              $\rightarrow$ bacterial mutation        &  83.6 & 5 \\
\hline
& M1  & war                       $\rightarrow$ argument                  &  93.5 & 7 \\
& M2  & buying an item            $\rightarrow$ accepting a belief        &  96.1 & 7 \\
& M3  & grounds for a building    $\rightarrow$ reasons for a theory      &  87.9 & 6 \\
& M4  & impediments to travel     $\rightarrow$ difficulties              & 100.0 & 7 \\
common
& M5  & money                     $\rightarrow$ time                      &  77.3 & 6 \\
metaphors
& M6  & seeds                     $\rightarrow$ ideas                     &  89.0 & 7 \\
& M7  & machine                   $\rightarrow$ mind                      &  98.7 & 7 \\
& M8  & object                    $\rightarrow$ idea                      &  89.1 & 5 \\
& M9  & following                 $\rightarrow$ understanding             &  96.6 & 8 \\
& M10 & seeing                    $\rightarrow$ understanding             &  78.8 & 6 \\
\hline
Average & & & 87.6 & 7.0 \\
\hline
\end{tabular}
\normalsize
\caption {The average agreement between our intended mappings and the
mappings of the 22 participants. See Appendix A for the details.}
\label{tab:agreement}
\end{table}

The column labeled $m$ gives the number of terms in the set of source terms
for each mapping problem (which is equal to the number of terms in the set of
target terms). For the average problem, $m = 7$.
The third column in Table~\ref{tab:agreement} gives a mnemonic that
summarizes the mapping (e.g., solar system $\rightarrow$ atom). Note that
the mnemonic is not used as input for any of the algorithms, nor
was the mnemonic shown to the participants in the experiment.

The agreement figures in Table~\ref{tab:agreement} for each individual
mapping problem are averages over the $m$ mappings for each problem.
Appendix A gives a more detailed view, showing the agreement for
each individual mapping in the $m$ mappings. The twenty problems contain
a total of 140 individual mappings ($20 \times 7$). Appendix A shows that
every one of these 140 mappings has an agreement of 50\% or higher. That is,
in every case, the majority of the participants agreed with our
intended mapping. (There are two cases where the agreement is
exactly 50\%. See problems A5 in Table~\ref{tab:a1-a5} and M5 in
Table~\ref{tab:m1-m5} in Appendix A.) 

If we select the mapping that is chosen by the majority of the 22 participants,
then we will get a perfect score on all twenty problems. More precisely,
if we try all $m!$ mappings for each problem, and select the mapping
that maximizes the sum of the number of participants who agree with
each individual mapping in the $m$ mappings, then we will have a
score of 100\% on all twenty problems. This is strong support for
the intended mappings that are given in Appendix A.

In Section~\ref{sec:task}, we applied Genter's \citeyear{gentner91} categories
-- {\em mere appearance} (mostly attributional similarity), {\em analogy} (mostly
relational similarity), and {\em literal similarity} (a mixture of attributional
and relational similarity) -- to the mappings $M_{\rm r}$ and $M_{\rm a}$,
where $M_{\rm r}$ is the best mapping according to ${\rm sim_r}$ and $M_{\rm a}$
is the best mapping according to ${\rm sim_a}$. The twenty mapping
problems were chosen as analogy problems; that is, the intended mappings
in Appendix A are meant to be relational mappings, $M_{\rm r}$; mappings that
maximize relational similarity, ${\rm sim_r}$. We have tried to avoid
mere appearance and literal similarity.

In Section~\ref{sec:lrme} we use the twenty mapping problems to evaluate
a relational mapping algorithm (LRME), and in Section~\ref{sec:attributes}
we use them to evaluate several different attributional mapping
algorithms. Our hypothesis is that LRME will perform significantly
better than any of the attributional mapping algorithms on the
twenty mapping problems, because they are analogy problems (not mere
appearance problems and not literal similarity problems).
We expect relational and attributional mapping algorithms
would perform approximately equally well on literal similarity problems,
and we expect that mere appearance problems would favour attributional
algorithms over relational algorithms, but we do not test these latter two
hypotheses, because our primary interest in this paper is analogy-making.

Our goal is to test the hypothesis that there is a
real, practical, effective, measurable difference between the output
of LRME and the output of the various attributional
mapping algorithms. A skeptic might claim that relational similarity
${\rm sim_r}(a\!:\!b, c\!:\!d)$ can be reduced to attributional
similarity ${\rm sim_a}(a,c) + {\rm sim_a}(b,d)$; therefore our
relational mapping algorithm is a complicated solution to an illusory
problem. A slightly less skeptical claim is that relational similarity
versus attributional similarity is a valid distinction in cognitive
psychology, but our relational mapping algorithm does not capture
this distinction. To test our hypothesis and refute these skeptical
claims, we have created twenty analogical mapping problems, and we
will show that LRME handles these problems significantly
better than the various attributional mapping algorithms.

%
%

\section{The Latent Relation Mapping Engine}
\label{sec:lrme}

The Latent Relation Mapping Engine (LRME) seeks the mapping $M_{\rm r}$
that maximizes the sum of the relational similarities.

\begin{equation}
\label{eqn:lrme}
M_{\rm r} = \operatornamewithlimits{arg\,max}_{M \in P(A,B)}
\; \sum_{i=1}^{m} \sum_{j=i+1}^{m} {\rm sim_r}(a_i\!:\!a_j, M(a_i)\!:\!M(a_j))
\end{equation}

\noindent We search for $M_{\rm r}$ by exhaustively evaluating all of the
possibilities. Ties are broken randomly. We use a simplified form of
LRA \shortcite{turney06} to calculate ${\rm sim_r}$.

\subsection{Algorithm}
\label{subsec:algorithm}

Briefly, the idea of LRME is to build a pair-pattern matrix $\mathbf{X}$,
in which the rows correspond to pairs of terms and the columns correspond
to patterns. For example, the row  $\mathbf{x}_{i:}$ might correspond to
the pair of terms {\em sun}$\,:\,${\em solar system} and the column
$\mathbf{x}_{:j}$ might correspond to the pattern ``$\ast$ $X$ centered $Y$ $\ast$''.
In these patterns, ``$\ast$'' is a wild card, which can match any single
word. The value of an element $x_{ij}$ in $\mathbf{X}$ is based on the frequency
of the pattern for $\mathbf{x}_{:j}$, when $X$ and $Y$ are instantiated by the
terms in the pair for $\mathbf{x}_{i:}$. For example, if we take the pattern
``$\ast$ $X$ centered $Y$ $\ast$'' and instantiate $X:Y$ with the pair
{\em sun}$\,:\,${\em solar system}, then we have the pattern
``$\ast$ sun centered solar system $\ast$'', and thus the value of the
element $x_{ij}$ is based on the frequency of ``$\ast$ sun centered solar system
$\ast$'' in the corpus. The matrix $\mathbf{X}$ is smoothed with a truncated
singular value decomposition (SVD) \shortcite{golub96} and the relational
similarity ${\rm sim_r}$ between two pairs of terms is given by the cosine
of the angle between the two corresponding row vectors in $\mathbf{X}$.

In more detail, LRME takes as input $I$ a set of mapping problems and
generates as output $O$ a corresponding set of mappings.

\begin{align}
I & = \left \{ \left \langle A_1, B_1 \right \rangle,
\left \langle A_2, B_2 \right \rangle, \ldots,
\left \langle A_n, B_n \right \rangle \right \} \\
O & = \left \{ M_1: A_1 \rightarrow B_1,
M_2: A_2 \rightarrow B_2, \ldots,
M_n: A_n \rightarrow B_n \right \}
\end{align}

\noindent In the following experiments, all twenty mapping problems
(Appendix A) are processed in one batch ($n = 20$).

The first step is to make a list $R$ that contains all pairs
of terms in the input $I$. For each mapping problem
$\left \langle A, B \right \rangle$ in $I$, we add to $R$ all pairs
$a_i:a_j$, such that $a_i$ and $a_j$ are members of $A$, $i \ne j$,
and all pairs
$b_i:b_j$, such that $b_i$ and $b_j$ are members of $B$, $i \ne j$.
If $\left | A \right | = \left | B \right | = m$, then there are
$m(m-1)$ pairs from $A$ and $m(m-1)$ pairs from $B$.\footnote{We have
$m(m-1)$ here, not $m(m-1)/2$, because we need the pairs in both orders.
We only want to calculate ${\rm sim_r}$ for one order of the pairs,
because $i$ is always less than $j$ in (\ref{eqn:lrme}); however, to ensure
that ${\rm sim_r}$ is symmetrical, as in (\ref{eqn:symmetry}), we need to
make the matrix $\mathbf{X}$ symmetrical, by having rows in the
matrix for both orders of every pair.}
A typical pair in $R$ would be {\em sun}$\,:\,${\em solar system}.
We do not allow duplicates in $R$; $R$ is a list of
pair types, not pair tokens. For our twenty mapping problems, $R$
is a list of 1,694 pairs.

For each pair $r$ in $R$, we make a list $S(r)$ of the phrases
in the corpus that contain the pair $r$. Let $a_i:a_j$ be the terms in the
pair $r$. We search in the corpus for all phrases of the following form:

\begin{equation}
\label{eqn:template}
\mbox{\textbf{``[0 to 1 words] $a_i$ [0 to 3 words] $a_j$ [0 to 1 words]''}}
\end{equation}

\noindent If $a_i:a_j$ is in $R$, then $a_j:a_i$ is also in $R$, so
we find phrases with the members of the pairs in both orders,
$S(a_i:a_j)$ and $S(a_j:a_i)$. The search template (\ref{eqn:template})
is the same as used by Turney \citeyear{turney08}.

In the following experiments, we search in a corpus of $5 \times 10^{10}$
English words (about 280 GB of plain text), consisting of web pages gathered
by a web crawler.\footnote{The corpus was collected by Charles Clarke at
the University of Waterloo. We can provide copies of the corpus on request.}
To retrieve phrases from the corpus, we use Wumpus \cite{buettcher05}, an
efficient search engine for passage retrieval
from large corpora.\footnote{Wumpus was developed by Stefan B{\"u}ttcher
and it is available at http://www.wumpus-search.org/.}

With the 1,694 pairs in $R$, we find a total of 1,996,464 phrases in the
corpus, an average of about 1,180 phrases per pair. For the pair
$r$ = {\em sun}$\,:\,${\em solar system}, a typical phrase $s$
in $S(r)$ would be ``a sun centered solar system illustrates''.

Next we make a list $C$ of patterns, based on the phrases we have found. 
For each pair $r$ in $R$, where $r = a_i:a_j$, if
we found a phrase $s$ in $S(r)$, then we replace $a_i$ in $s$ with $X$
and we replace $a_j$ with $Y$. The remaining words may
be either left as they are or replaced with a wild card symbol ``$\ast$''.
We then replace $a_i$ in $s$ with $Y$ and $a_j$ with $X$, and
replace the remaining words with wild cards or leave them as
they are. If there are $n$ remaining words in $s$, after $a_i$
and $a_j$ are replaced, then we generate $2^{n+1}$ patterns from $s$,
and we add these patterns to $C$. We only add new patterns to $C$;
that is, $C$ is a list of pattern types, not pattern tokens; there
are no duplicates in $C$.

For example, for the pair {\em sun}$\,:\,${\em solar system},
we found the phrase ``a sun centered solar system illustrates''.
When we replace $a_i:a_j$ with $X:Y$, we have
``a $X$ centered $Y$ illustrates''. There are three remaining words,
so we can generate eight patterns, such as ``a $X$ $\ast$ $Y$ illustrates'',
``a $X$ centered $Y$ $\ast$'', ``$\ast$ $X$ $\ast$ $Y$ illustrates'', and so on.
Each of these patterns is added to $C$. Then we replace $a_i:a_j$
with $Y:X$, yielding ``a $Y$ centered $X$ illustrates''. This
gives us another eight patterns, such as ``a $Y$ centered $X$ $\ast$''.
Thus the phrase ``a sun centered solar system illustrates'' generates
a total of sixteen patterns, which we add to $C$.

Now we revise $R$, to make a list of pairs that will correspond to
rows in the frequency matrix $\mathbf{F}$. We remove any pairs from $R$
for which no phrases were found in the corpus, when the terms were
in either order. Let $a_i:a_j$ be the terms in the pair $r$.
We remove $r$ from $R$ if both $S(a_i:a_j)$ and $S(a_j:a_i)$ are empty.
We remove such rows because they would correspond
to zero vectors in the matrix $\mathbf{F}$. This reduces $R$ from 1,694
pairs to 1,662 pairs. Let $n_r$ be the number of pairs in $R$.

Next we revise $C$, to make a list of patterns that will correspond
to columns in the frequency matrix $\mathbf{F}$. In the following
experiments, at this stage, $C$ contains millions of patterns, too many for
efficient processing with a standard desktop computer. We need
to reduce $C$ to a more manageable size. We select the patterns that
are shared by the most pairs.
Let $c$ be a pattern in $C$. Let $r$ be a pair in $R$.
If there is a phrase $s$ in $S(r)$, such that there is a pattern generated
from $s$ that is identical to $c$, then we say that $r$ is one of the
pairs that generated $c$. We sort the patterns in $C$ in descending order of the
number of pairs in $R$ that generated each pattern, and we select the top
$tn_r$ patterns from this sorted list. Following Turney \citeyear{turney08},
we set the parameter $t$ to 20; hence $C$ is reduced to the top 33,240
patterns ($tn_r$ = 20 $\times$ 1,662 = 33,240). Let $n_c$ be the
number of patterns in $C$ ($n_c = tn_r)$.

Now that the rows $R$ and columns $C$ are defined, we can build
the frequency matrix $\mathbf{F}$.
Let $r_i$ be the $i$-th pair of terms in $R$ (e.g., let $r_i$ be
{\em sun}$\,:\,${\em solar system}) and let $c_j$ be the $j$-th pattern
in $C$ (e.g., let $c_j$ be ``$\ast$ $X$ centered $Y$ $\ast$'').
We instantiate $X$ and $Y$ in the pattern $c_j$ with the
terms in $r_i$ (``$\ast$ sun centered solar system $\ast$'').
The element $f_{ij}$ in $\mathbf{F}$ is the frequency
of this instantiated pattern in the corpus.

Note that we do not need to search again in the corpus for the instantiated
pattern for $f_{ij}$, in order to find its frequency. In the process of creating
each pattern, we can keep track of how many phrases generated the
pattern, for each pair. We can get the frequency for $f_{ij}$ by
checking our record of the patterns that were generated by $r_i$.

The next step is to transform the matrix $\mathbf{F}$ of raw frequencies
into a form $\mathbf{X}$ that enhances the similarity measurement. Turney
\citeyear{turney06} used the log entropy transformation, as suggested
by Landauer and Dumais \citeyear{landauer97}. This is a kind of
tf-idf (term frequency times inverse document frequency) transformation,
which gives more weight to elements in the matrix that are statistically
surprising. However, Bullinaria and Levy \citeyear{bullinaria07} recently
achieved good results with a new transformation, called PPMIC
(Positive Pointwise Mutual Information with Cosine); therefore
LRME uses PPMIC. The raw frequencies in $\mathbf{F}$ are used to calculate
probabilities, from which we can calculate the pointwise mutual information
(PMI) of each element in the matrix. Any element with a negative PMI is then
set to zero.

\begin{align}
p_{ij} & = \frac{f_{ij}}{\sum_{i=1}^{n_r} \sum_{j=1}^{n_c} f_{ij}} \\
p_{i*} & = \frac{\sum_{j=1}^{n_c} f_{ij}}{\sum_{i=1}^{n_r} \sum_{j=1}^{n_c} f_{ij}} \\
p_{*j} & = \frac{\sum_{i=1}^{n_r} f_{ij}}{\sum_{i=1}^{n_r} \sum_{j=1}^{n_c} f_{ij}} \\
\label{eqn:pmi}
{\rm pmi}_{ij} & = \log \left ( \frac{p_{ij}}{p_{i*} p_{*j}} \right ) \\
x_{ij} & =
\left\{
\begin{array}{rl}
{\rm pmi}_{ij} & \mbox{if ${\rm pmi}_{ij} > 0$} \\
0 & \mbox{otherwise}
\end{array}
\right.
\end{align}

Let $r_i$ be the \mbox{$i$-th} pair of terms in $R$ (e.g., let $r_i$ be
{\em sun}$\,:\,${\em solar system}) and let $c_j$ be the \mbox{$j$-th} pattern
in $C$ (e.g., let $c_j$ be ``$\ast$ $X$ centered $Y$ $\ast$'').
In (\ref{eqn:pmi}), $p_{ij}$ is the estimated probability
of the of the pattern $c_j$ instantiated with the pair $r_i$
(``$\ast$ sun centered solar system $\ast$''), $p_{i*}$
is the estimated probability of $r_i$, and $p_{*j}$ is
the estimated probability of $c_j$. If $r_i$ and $c_j$ are
statistically independent, then $p_{i*} p_{*j} = p_{ij}$ (by the
definition of independence), and
thus ${\rm pmi}_{ij}$ is zero (since $\log(1) = 0$). If there is an
interesting semantic relation between the terms in $r_i$, and the
pattern $c_j$ captures an aspect of that semantic relation, then
we should expect $p_{ij}$ to be larger than it would be if
$r_i$ and $c_j$ were indepedent; hence we should find that
$p_{ij} > p_{i*} p_{*j}$, and thus ${\rm pmi}_{ij}$ is positive.
(See Hypothesis~2 in Section~\ref{sec:hypotheses}.) On the
other hand, terms from completely different domains may avoid
each other, in which case we should find that ${\rm pmi}_{ij}$
is negative. PPMIC is designed to give a high value to
$x_{ij}$ when the pattern $c_j$ captures an aspect of the
semantic relation between the terms in $r_i$; otherwise,
$x_{ij}$ should have a value of zero, indicating that the
pattern $c_j$ tells us nothing about the semantic relation
between the terms in $r_i$.

In our experiments, $\mathbf{F}$ has a density of 4.6\% (the
percentage of nonzero elements) and
$\mathbf{X}$ has a density of 3.8\%. The lower density of $\mathbf{X}$
is due to elements with a negative PMI, which are transformed to zero
by PPMIC.

Now we smooth $\mathbf{X}$ by applying a truncated singular value decomposition
(SVD) \shortcite{golub96}. We use SVDLIBC to calculate the SVD of
$\mathbf{X}$.\footnote{SVDLIBC is the work of Doug Rohde and it is available
at http://tedlab.mit.edu/$\scriptstyle\sim$dr/svdlibc/.}
SVDLIBC is designed for sparse (low density) matrices.
SVD decomposes $\mathbf{X}$ into the product of three matrices
$\mathbf{U} \mathbf{\Sigma} \mathbf{V}^\mathsf{T}$, where $\mathbf{U}$
and $\mathbf{V}$ are in column
orthonormal form (i.e., the columns are orthogonal and have unit length,
$\mathbf{U}^\mathsf{T} \mathbf{U} = \mathbf{V}^\mathsf{T} \mathbf{V} = \mathbf{I}$)
and $\mathbf{\Sigma}$ is a diagonal matrix of singular values \shortcite{golub96}.
If $\mathbf{X}$ is of rank $r$, then $\mathbf{\Sigma}$ is also of rank $r$.
Let ${\mathbf{\Sigma}}_k$, where $k < r$, be the diagonal matrix formed from the top $k$
singular values, and let $\mathbf{U}_k$ and $\mathbf{V}_k$ be the matrices produced
by selecting the corresponding columns from $\mathbf{U}$ and $\mathbf{V}$. The matrix
$\mathbf{U}_k \mathbf{\Sigma}_k \mathbf{V}_k^\mathsf{T}$ is the matrix of rank $k$
that best approximates the original matrix $\mathbf{X}$, in the sense that it
minimizes the approximation errors. That is,
${\bf \hat X} = \mathbf{U}_k \mathbf{\Sigma}_k \mathbf{V}_k^\mathsf{T}$
minimizes $\| {{\bf \hat X} - \mathbf{X}} \|_F$
over all matrices ${\bf \hat X}$ of rank $k$, where $\| \ldots \|_F$
denotes the Frobenius norm \shortcite{golub96}. We may think of this matrix
$\mathbf{U}_k \mathbf{\Sigma}_k \mathbf{V}_k^\mathsf{T}$ as a smoothed or compressed
version of the original matrix $\mathbf{X}$. Following Turney \citeyear{turney06},
we set the parameter $k$ to 300.

The relational similarity ${\rm sim_r}$ between two pairs in $R$ is
the inner product of the two corresponding rows in
$\mathbf{U}_k \mathbf{\Sigma}_k \mathbf{V}_k^\mathsf{T}$,
after the rows have been normalized to unit length. We can simplify
calculations by dropping $\mathbf{V}_k$ \cite{deerwester90}.
We take the matrix $\mathbf{U}_k \mathbf{\Sigma}_k$ and normalize
each row to unit length. Let $\mathbf{W}$ be the resulting
matrix. Now let $\mathbf{Z}$ be $\mathbf{W} \mathbf{W}^\mathsf{T}$,
a square matrix of size $n_r \times n_r$. This matrix contains the
cosines of all combinations of two pairs in $R$.

For a mapping problem $\left \langle A, B \right \rangle$ in $I$,
let $a:a'$ be a pair of terms from $A$ and let $b:b'$ be a pair
of terms from $B$. Suppose that $r_i = a:a'$ and $r_j = b:b'$,
where $r_i$ and $r_j$ are the $i$-th and $j$-th pairs in $R$. Then
${\rm sim_r}(a:a', b:b') = z_{ij}$, where $z_{ij}$ is the
element in the $i$-th row and $j$-th column of $\mathbf{Z}$.
If either $a:a'$ or $b:b'$ is not in $R$, because $S(a:a')$,
$S(a':a)$, $S(b:b')$, or $S(b':b)$ is empty, then we set the
similarity to zero. Finally, for each mapping problem in $I$,
we output the map $M_{\rm r}$ that maximizes the sum of the relational
similarities.

\begin{equation}
M_{\rm r} = \operatornamewithlimits{arg\,max}_{M \in P(A,B)}
\; \sum_{i=1}^{m} \sum_{j=i+1}^{m} {\rm sim_r}(a_i\!:\!a_j, M(a_i)\!:\!M(a_j))
\end{equation}

The simplified form of LRA used here to calculate ${\rm sim_r}$
differs from LRA used by Turney \citeyear{turney06} in several ways.
In LRME, there is no use of synonyms to generate alternate forms of
the pairs of terms. In LRME, there is no morphological processing of the terms.
LRME uses PPMIC \shortcite{bullinaria07} to process the raw frequencies,
instead of log entropy. Following Turney \citeyear{turney08}, LRME uses a
slightly different search template (\ref{eqn:template}) and LRME sets the
number of columns $n_c$ to $tn_r$, instead of using a constant. In
Section~\ref{subsec:experiments}, we evaluate the impact of two of these
changes (PPMIC and $n_c$), but we have not tested the other changes, which
were mainly motivated by a desire for increased efficiency and simplicity.

\subsection{Experiments}
\label{subsec:experiments}

We implemented LRME in Perl, making external calls
to Wumpus for searching the corpus and to SVDLIBC for calculating SVD.
We used the Perl Net::Telnet package for interprocess communication
with Wumpus, the PDL (Perl Data Language) package for matrix manipulations
(e.g., calculating cosines), and the List::Permutor package to
generate permutations (i.e., to loop through $P(A,B)$).

We ran the following experiments on a dual core AMD Opteron 64
computer, running 64 bit Linux. Most of the running time is spent
searching the corpus for phrases. It took 16 hours
and 27 minutes for Wumpus to fetch the 1,996,464 phrases.
The remaining steps took 52 minutes, of which SVD took 10
minutes. The running time could be cut in half by using
RAID 0 to speed up disk access.

Table~\ref{tab:lrme-baseline} shows the performance of LRME in its baseline
configuration. For comparison, the agreement of the 22 volunteers with
our intended mapping has been copied from Table~\ref{tab:agreement}.
The difference between the performance of LRME (91.5\%) and the human participants
(87.6\%) is not statistically significant (paired t-test, 95\% confidence level).

\begin{table}[htbp]
\small
\centering
\begin{tabular}{llrr}
\hline
& & \multicolumn{2}{c}{\textbf{Accuracy}} \\
\cline{3-4}
\textbf{Mapping} & \textbf{Source $\rightarrow$ Target} & \textbf{LRME} & \textbf{Humans} \\
\hline
A1  & solar system            $\rightarrow$ atom                  &  100.0 &  90.9 \\
A2  & water flow              $\rightarrow$ heat transfer         &  100.0 &  86.9 \\
A3  & waves                   $\rightarrow$ sounds                &  100.0 &  81.8 \\
A4  & combustion              $\rightarrow$ respiration           &  100.0 &  79.0 \\
A5  & sound                   $\rightarrow$ light                 &   71.4 &  79.2 \\
A6  & projectile              $\rightarrow$ planet                &  100.0 &  97.4 \\
A7  & artificial selection    $\rightarrow$ natural selection     &   71.4 &  74.7 \\
A8  & billiard balls          $\rightarrow$ gas molecules         &  100.0 &  88.1 \\
A9  & computer                $\rightarrow$ mind                  &   55.6 &  84.3 \\
A10 & slot machine            $\rightarrow$ bacterial mutation    &  100.0 &  83.6 \\
\hline
M1  & war                     $\rightarrow$ argument              &   71.4 &  93.5 \\
M2  & buying an item          $\rightarrow$ accepting a belief    &  100.0 &  96.1 \\
M3  & grounds for a building  $\rightarrow$ reasons for a theory  &  100.0 &  87.9 \\
M4  & impediments to travel   $\rightarrow$ difficulties          &  100.0 & 100.0 \\
M5  & money                   $\rightarrow$ time                  &  100.0 &  77.3 \\
M6  & seeds                   $\rightarrow$ ideas                 &  100.0 &  89.0 \\
M7  & machine                 $\rightarrow$ mind                  &  100.0 &  98.7 \\
M8  & object                  $\rightarrow$ idea                  &   60.0 &  89.1 \\
M9  & following               $\rightarrow$ understanding         &  100.0 &  96.6 \\
M10 & seeing                  $\rightarrow$ understanding         &  100.0 &  78.8 \\
\hline
Average & & 91.5 & 87.6 \\
\hline
\end{tabular}
\normalsize
\caption {LRME in its baseline configuration, compared with human performance.}
\label{tab:lrme-baseline}
\end{table}

In Table~\ref{tab:lrme-baseline}, the column labeled {\em Humans} is the average
of 22 people, whereas the {\em LRME} column is only one algorithm (it is not
an average). Comparing an average of several scores to an individual score
(whether the individual is a human or an algorithm) may give a misleading
impression. In the results for any individual
person, there are typically several 100\% scores and a few scores in the 55-75\%
range. The average mapping problem has seven terms. It is not possible to have
exactly one term mapped incorrectly; if there are any incorrect mappings,
then there must be two or more incorrect mappings. This follows from the
nature of bijections. Therefore a score of $5/7 = 71.4\%$ is not uncommon.

Table~\ref{tab:lrme-histogram} looks at the results from another
perspective. The column labeled {\em LRME wrong} gives the number
of incorrect mappings made by LRME for each of the twenty problems.
The five columns labeled {\em Number of people with $N$ wrong} show,
for various values of $N$, how may of the 22 people made $N$ incorrect
mappings. For the average mapping problem, 15 out of 22 participants had a perfect
score ($N = 0$); of the remaining 7 participants, 5 made only two mistakes
($N = 2$). Table~\ref{tab:lrme-histogram} shows more clearly than
Table~\ref{tab:lrme-baseline} that LRME's performance is not significantly
different from (individual) human performance. (For yet another perspective,
see Section~\ref{subsec:analogies-vs-metaphors}).

\begin{table}[htbp]
\small
\centering
\begin{tabular}{lccccccc}
\hline
& \textbf{LRME} & \multicolumn{5}{c}{\textbf{Number of people with $N$ wrong}} & \\
\cline{3-7}
\textbf{Mapping} & \textbf{wrong} & \textbf{$N = 0$}
& \textbf{$N = 1$} & \textbf{$N = 2$}
& \textbf{$N = 3$} & \textbf{$N \ge 4$} & \textbf{$m$} \\
\hline
A1            &   0  &  16 &  0 &  4 &  2 &  0 &  7 \\
A2            &   0  &  14 &  0 &  5 &  0 &  3 &  8 \\
A3            &   0  &   9 &  0 &  9 &  2 &  2 &  8 \\
A4            &   0  &   9 &  0 &  9 &  0 &  4 &  8 \\
A5            &   2  &  10 &  0 &  7 &  2 &  3 &  7 \\
A6            &   0  &  20 &  0 &  2 &  0 &  0 &  7 \\
A7            &   2  &   8 &  0 &  6 &  6 &  2 &  7 \\
A8            &   0  &  13 &  0 &  8 &  0 &  1 &  8 \\
A9            &   4  &  11 &  0 &  7 &  2 &  2 &  9 \\
A10           &   0  &  13 &  0 &  9 &  0 &  0 &  5 \\
\hline
M1            &   2  &  17 &  0 &  5 &  0 &  0 &  7 \\
M2            &   0  &  19 &  0 &  3 &  0 &  0 &  7 \\
M3            &   0  &  14 &  0 &  8 &  0 &  0 &  6 \\
M4            &   0  &  22 &  0 &  0 &  0 &  0 &  7 \\
M5            &   0  &   9 &  0 & 11 &  0 &  2 &  6 \\
M6            &   0  &  15 &  0 &  4 &  3 &  0 &  7 \\
M7            &   0  &  21 &  0 &  1 &  0 &  0 &  7 \\
M8            &   2  &  18 &  0 &  2 &  1 &  1 &  5 \\
M9            &   0  &  19 &  0 &  3 &  0 &  0 &  8 \\
M10           &   0  &  13 &  0 &  3 &  3 &  3 &  6 \\
\hline
Average       &   1  &  15 &  0 &  5 &  1 &  1 &  7 \\
\hline
\end{tabular}
\normalsize
\caption {Another way of viewing LRME versus human performance.}
\label{tab:lrme-histogram}
\end{table}

In Table~\ref{tab:lrme-variations}, we examine the sensitivity of LRME
to the parameter settings. The first row shows the accuracy of the
baseline configuration, as in Table~\ref{tab:lrme-baseline}. The next
eight rows show the impact of varying $k$, the dimensionality of the
truncated singular value decomposition, from 50 to 400. The eight rows
after that show the effect of varying $t$, the column factor, from
5 to 40. The number of columns in the matrix ($n_c$) is given by
the number of rows ($n_r$ = 1,662) multiplied by $t$. The second last row
shows the effect of eliminating the singular value decomposition
from LRME. This is equivalent to setting $k$ to 1,662, the number of
rows in the matrix. The final row gives the result when PPMIC
\shortcite{bullinaria07} is replaced with log entropy \shortcite{turney06}.
LRME is not sensitive to any of these manipulations: None
of the variations in Table~\ref{tab:lrme-variations} perform
significantly differently from the baseline configuration
(paired t-test, 95\% confidence level). (This does not necessarily
mean that the manipulations have no effect; rather, it suggests
that a larger sample of problems would be needed to show
a significant effect.)

\begin{table}[htbp]
\small
\centering
\begin{tabular}{lrrrr}
\hline
\textbf{Experiment} & \textbf{$k$} & \textbf{$t$} & \textbf{$n_c$} & \textbf{Accuracy} \\
\hline
baseline configuration
&  300   &    20    &    33,240  &   91.5 \\
\hline
\multirow{8}{*}{varying $k$}
&   50   &    20    &    33,240  &   89.3 \\
&  100   &    20    &    33,240  &   92.8 \\
&  150   &    20    &    33,240  &   91.3 \\
&  200   &    20    &    33,240  &   92.6 \\
&  250   &    20    &    33,240  &   90.6 \\
&  300   &    20    &    33,240  &   91.5 \\
&  350   &    20    &    33,240  &   90.6 \\
&  400   &    20    &    33,240  &   90.6 \\
\hline
\multirow{8}{*}{varying $t$}
&  300   &     5    &     8,310  &   86.9 \\
&  300   &    10    &    16,620  &   94.0 \\
&  300   &    15    &    24,930  &   94.0 \\
&  300   &    20    &    33,240  &   91.5 \\
&  300   &    25    &    41,550  &   90.1 \\
&  300   &    30    &    49,860  &   90.6 \\
&  300   &    35    &    58,170  &   89.5 \\
&  300   &    40    &    66,480  &   91.7 \\
\hline
dropping SVD
& 1662   &    20    &    33,240  &   89.7 \\
\hline
log entropy
&  300   &    20    &    33,240  &   83.9 \\
\hline
\end{tabular}
\normalsize
\caption {Exploring the sensitivity of LRME to various
parameter settings and modifications.}
\label{tab:lrme-variations}
\end{table}

%
%

\section{Attribute Mapping Approaches}
\label{sec:attributes}

In this section, we explore a variety of attribute mapping approaches
for the twenty mapping problems. All of these approaches seek the
mapping $M_{\rm a}$ that maximizes the sum of the attributional similarities.

\begin{equation}
M_{\rm a} = \operatornamewithlimits{arg\,max}_{M \in P(A,B)}
\; \sum_{i=1}^{m} {\rm sim_a}(a_i, M(a_i))
\end{equation}

\noindent We search for $M_{\rm a}$ by exhaustively evaluating all of the
possibilities. Ties are broken randomly. We use a variety of different
algorithms to calculate ${\rm sim_a}$.

\subsection{Algorithms}
\label{subsec:attrib-algo}

In the following experiments, we test five lexicon-based attributional
similarity measures that use
WordNet:\footnote{WordNet was developed by a team at Princeton
and it is available at http://wordnet.princeton.edu/.}
HSO \shortcite{hirst98}, JC \shortcite{jiang97},
LC \shortcite{leacock98}, LIN \shortcite{lin98}, and
RES \shortcite{resnik95}. All five are implemented in the Perl package
WordNet::Similarity,\footnote{Ted Pedersen's WordNet::Similarity
package is at
http://www.d.umn.edu/$\scriptstyle\sim$tpederse/similarity.html.}
which builds on the
WordNet::QueryData\footnote{Jason Rennie's WordNet::QueryData
package is at
http://people.csail.mit.edu/jrennie/WordNet/.}
package. The core idea behind them is to treat WordNet as
a graph and measure the semantic distance between two terms by
the length of the shortest path between them in the graph.
Similarity increases as distance decreases.

HSO works with nouns, verbs, adjectives, and adverbs, but JC, LC, LIN,
and RES only work with nouns and verbs. We used WordNet::Similarity
to try all possible parts of speech and all possible senses for each
input word. Many adjectives, such as {\em true} and {\em valuable}, also
have noun and verb senses in WordNet, so JC, LC, LIN, and RES are still
able to calculate similarity for them. When the raw form of a word is not
found in WordNet, WordNet::Similarity searches for morphological variations
of the word. When there are multiple similarity scores, for
multiple parts of speech and multiple senses, we select the highest
similarity score. When there is no similarity score, because a word
is not in WordNet, or because JC, LC, LIN, or RES could not find
an alternative noun or verb form for an adjective or adverb, we
set the score to zero.

We also evaluate two corpus-based attributional similarity measures:
PMI-IR \shortcite{turney01} and LSA \shortcite{landauer97}.
The core idea behind them is that ``a word is characterized
by the company it keeps'' \shortcite{firth57}.
The similarity of two terms is measured by the similarity
of their statistical distributions in a corpus.
We used the corpus of Section~\ref{sec:lrme} along with
Wumpus to implement PMI-IR (Pointwise Mutual Information with
Information Retrieval). For LSA (Latent Semantic Analysis), we used
the online demonstration.\footnote{The online demonstration of LSA
is the work of a team at the University of Colorado at Boulder.
It is available at http://lsa.colorado.edu/.} We selected the
{\em Matrix Comparison} option with the {\em General Reading up to
1st year college (300 factors)} topic space and the {\em term-to-term}
comparison type. PMI-IR and LSA work with all parts of speech.

Our eighth similarity measure is based on the observation that our
intended mappings map terms that have the same part of speech (see
Appendix A). Let ${\rm POS}(a)$ be the part-of-speech tag assigned
to the term $a$. We use part-of-speech tags to define a measure of
attributional similarity,
${\rm sim_{\scriptscriptstyle POS}}(a, b)$, as follows.

\begin{equation}
\label{eqn:pos}
{\rm sim_{\scriptscriptstyle POS}}(a, b) =
\left\{
\begin{array}{rl}
100 & \mbox{if $a = b$} \\
10  & \mbox{if ${\rm POS}(a) = {\rm POS}(b)$} \\
0   & \mbox{otherwise}
\end{array}
\right.
\end{equation}

\noindent We hand-labeled the terms in the mapping problems
with part-of-speech tags \shortcite{santorini90}. Automatic
taggers assume that the words that are to be tagged are embedded
in a sentence, but the terms in our mapping problems are not
in sentences, so their tags are ambiguous. We used our
knowledge of the intended mappings to manually disambiguate the
part-of-speech tags for the terms, thus guaranteeing that corresponding
terms in the intended mapping always have the same tags.

For each of the first seven attributional similarity measures
above, we created seven more similarity measures by combining
them with ${\rm sim_{\scriptscriptstyle POS}}(a, b)$. For example,
let ${\rm sim_{\scriptscriptstyle HSO}}(a, b)$ be the Hirst and
St-Onge \citeyear{hirst98} similarity measure. We combine
${\rm sim_{\scriptscriptstyle POS}}(a, b)$ and
${\rm sim_{\scriptscriptstyle HSO}}(a, b)$ by simply adding them.

\begin{equation}
{\rm sim_{\scriptscriptstyle HSO+POS}}(a, b) =
{\rm sim_{\scriptscriptstyle HSO}}(a, b) +
{\rm sim_{\scriptscriptstyle POS}}(a, b)
\end{equation}

\noindent The values returned by ${\rm sim_{\scriptscriptstyle POS}}(a, b)$
range from 0 to 100, whereas the values returned by
${\rm sim_{\scriptscriptstyle HSO}}(a, b)$ are
much smaller. We chose large values in (\ref{eqn:pos}) so that
getting POS tags to match up has more weight than any of the
other similarity measures. The manual POS
tags and the high weight of
${\rm sim_{\scriptscriptstyle POS}}(a, b)$
give an unfair advantage to the attributional mapping approach,
but the relational mapping approach can afford to be generous.

\subsection{Experiments}
\label{subsec:attrib-exper}

Table~\ref{tab:attributional} presents the accuracy of the
various measures of attributional similarity. The best
result without POS labels is 55.9\% (HSO). The best result with POS
labels is 76.8\% (LIN+POS). The 91.5\% accuracy of LRME
(see Table~\ref{tab:lrme-baseline}) is significantly higher than
the 76.8\% accuracy of LIN+POS (and thus, of course, significantly
higher than everything else in Table~\ref{tab:attributional};
paired t-test, 95\% confidence level). The average human performance
of 87.6\% (see Table~\ref{tab:agreement}) is also significantly higher
than the 76.8\% accuracy of LIN+POS (paired \mbox{t-test}, 95\% confidence
level). In summary, humans and LRME perform significantly better than all
of the variations of attributional mapping approaches that were tested.

\begin{table}[htbp]
\small
\centering
\begin{tabular}{llr}
\hline
\textbf{Algorithm} & \textbf{Reference} & \textbf{Accuracy} \\
\hline
HSO                & Hirst and St-Onge \citeyear{hirst98}       & 55.9 \\
JC                 & Jiang and Conrath \citeyear{jiang97}       & 54.7 \\
LC                 & Leacock and Chodrow \citeyear{leacock98}   & 48.5 \\
LIN                & Lin \citeyear{lin98}                       & 48.2 \\
RES                & Resnik \citeyear{resnik95}                 & 43.8 \\
PMI-IR             & Turney \citeyear{turney01}                 & 54.4 \\
LSA                & Landauer and Dumais \citeyear{landauer97}  & 39.6 \\
\hline
POS (hand-labeled) & Santorini \citeyear{santorini90}           & 44.8 \\
\hline
HSO+POS            & Hirst and St-Onge \citeyear{hirst98}       & 71.1 \\
JC+POS             & Jiang and Conrath \citeyear{jiang97}       & 73.6 \\
LC+POS             & Leacock and Chodrow \citeyear{leacock98}   & 69.5 \\
LIN+POS            & Lin \citeyear{lin98}                       & 76.8 \\
RES+POS            & Resnik \citeyear{resnik95}                 & 71.6 \\
PMI-IR+POS         & Turney \citeyear{turney01}                 & 72.8 \\
LSA+POS            & Landauer and Dumais \citeyear{landauer97}  & 65.8 \\
\hline
\end{tabular}
\normalsize
\caption {The accuracy of attribute mapping approaches for a wide
variety of measures of attributional similarity.}
\label{tab:attributional}
\end{table}

%
%

\section{Discussion}
\label{sec:discussion}

In this section, we examine three questions that are suggested by the
preceding results. Is there a difference between the science analogy
problems and the common metaphor problems? Is there an advantage
to combining the relational and attributional mapping approaches?
What is the advantage of the relational mapping approach over the
attributional mapping approach?

\subsection{Science Analogies versus Common Metaphors}
\label{subsec:analogies-vs-metaphors}

Table~\ref{tab:agreement} suggests that science analogies may be
more difficult than common metaphors. This is supported by
Table~\ref{tab:sci-met-human}, which shows how the agreement of
the 22 participants with our intended mapping (see Section~\ref{sec:problems})
varies between the science problems and the metaphor problems.
The science problems have a lower average performance and greater
variation in performance. The difference between the science problems
and the metaphor problems is statistically significant (paired t-test,
95\% confidence level).

\begin{table}[htbp]
\small
\centering
\begin{tabular}{lrrr}
\hline
& \multicolumn{3}{c}{\textbf{Average Accuracy}} \\
\cline{2-4}
\textbf{Participant} & \textbf{All 20}
& \textbf{10 Science} & \textbf{10 Metaphor} \\
\hline
1        &          72.6   &          59.9   &           85.4  \\
2        &          88.2   &          85.9   &           90.5  \\
3        &          90.0   &          86.3   &   \textbf{93.8} \\
4        &          71.8   &          56.4   &           87.1  \\
5        &  \textbf{95.7}  &  \textbf{94.2}  &   \textbf{97.1} \\
6        &          83.4   &          83.9   &           82.9  \\
7        &          79.6   &          73.6   &           85.7  \\
8        &  \textbf{91.9}  &  \textbf{95.0}  &           88.8  \\
9        &          89.7   &  \textbf{90.0}  &           89.3  \\
10       &          80.7   &          81.4   &           80.0  \\
11       &  \textbf{94.5}  &  \textbf{95.7}  &   \textbf{93.3} \\
12       &          90.6   &          87.4   &   \textbf{93.8} \\
13       &  \textbf{93.2}  &          89.6   &   \textbf{96.7} \\
14       &  \textbf{97.1}  &  \textbf{94.3}  &  \textbf{100.0} \\
15       &          86.6   &          88.5   &           84.8  \\
16       &          80.5   &          80.2   &           80.7  \\
17       &  \textbf{93.3}  &  \textbf{89.9}  &   \textbf{96.7} \\
18       &          86.5   &          78.9   &   \textbf{94.2} \\
19       &  \textbf{92.9}  &  \textbf{96.0}  &           89.8  \\
20       &          90.4   &          84.1   &   \textbf{96.7} \\
21       &          82.7   &          74.9   &           90.5  \\
22       &  \textbf{96.2}  &  \textbf{94.9}  &   \textbf{97.5} \\
\hline
Average  &  87.6  &  84.6  &   90.7 \\
Standard deviation  &   7.2  &  10.8  &    5.8 \\
\hline
\end{tabular}
\normalsize
\caption {A comparison of the difficulty of the science problems
versus the metaphor problems for the 22 participants. The numbers
in bold font are the scores that are above the scores of LRME.}
\label{tab:sci-met-human}
\end{table}

The average science problem has more terms (7.4) than the average
metaphor problem (6.6), which might contribute to the difficulty
of the science problems. However, Table~\ref{tab:terms} shows that
there is no clear relation between the number of terms in a problem
($m$ in Table~\ref{tab:agreement}) and the level of agreement.
We believe that people find the metaphor problems easier than the
science problems because these common metaphors are entrenched
in our language, whereas the science analogies are more
peripheral.

\begin{table}[htbp]
\normalsize
\centering
\begin{tabular}{cc}
\hline
\textbf{Num terms} & \textbf{Agreement} \\
\hline
5 & 86.4 \\
6 & 81.3 \\
7 & 91.1 \\
8 & 86.5 \\
9 & 84.3 \\
\hline
\end{tabular}
\normalsize
\caption {The average agreement among the 22 participants as a
function of the number of terms in the problems.}
\label{tab:terms}
\end{table}

Table~\ref{tab:sci-met-alg} shows that the 16 algorithms studied
here perform slightly worse on the science problems than on the
metaphor problems, but the difference is not statistically
significant (paired t-test, 95\% confidence level). We hypothesize
that the attributional mapping approaches are not performing
well enough to be sensitive to subtle differences between
science analogies and common metaphors.

\begin{table}[htb]
\small
\centering
\begin{tabular}{lrrr}
\hline
& \multicolumn{3}{c}{\textbf{Average Accuracy}} \\
\cline{2-4}
\textbf{Algorithm} & \textbf{All 20} & \textbf{10 Science} & \textbf{10 Metaphor} \\
\hline
LRME       &  91.5  &  89.8  &  93.1 \\
\hline
HSO        &  55.9  &  57.4  &  54.3 \\
JC         &  54.7  &  57.4  &  52.1 \\
LC         &  48.5  &  49.6  &  47.5 \\
LIN        &  48.2  &  46.7  &  49.7 \\
RES        &  43.8  &  39.0  &  48.6 \\
PMI-IR     &  54.4  &  49.5  &  59.2 \\
LSA        &  39.6  &  37.3  &  41.9 \\
\hline
POS        &  44.8  &  42.1  &  47.4 \\
\hline
HSO+POS    &  71.1  &  66.9  &  75.2 \\
JC+POS     &  73.6  &  78.1  &  69.2 \\
LC+POS     &  69.5  &  70.8  &  68.2 \\
LIN+POS    &  76.8  &  68.8  &  84.8 \\
RES+POS    &  71.6  &  70.3  &  72.9 \\
PMI-IR+POS &  72.8  &  65.7  &  79.9 \\
LSA+POS    &  65.8  &  69.1  &  62.4 \\
\hline
Average    &  61.4  &  59.9  &  62.9 \\
Standard deviation  &  14.7  &  15.0  &  15.3 \\
\hline
\end{tabular}
\normalsize
\caption {A comparison of the difficulty of the science problems
versus the metaphor problems for the 16 algorithms.}
\label{tab:sci-met-alg}
\end{table}

Incidentally, these tables give us another view of the performance
of LRME in comparison to human performance. The first row in
Table~\ref{tab:sci-met-alg} shows the performance of LRME on the
science and metaphor problems. In Table~\ref{tab:sci-met-human},
we have marked in bold font the cases where human scores are
greater than LRME's scores. For all 20 problems, there
are 8 such cases; for the 10 science problems, there are 8
such cases; for the 10 metaphor problems, there are 10 such
cases. This is further evidence that LRME's performance is
not significantly different from human performance. LRME is near
the middle of the range of performance of the 22 human participants.

\subsection{Hybrid Relational-Attributional Approaches}
\label{subsec:hybrids}

Recall the definitions of ${\rm score_r}(M)$ and ${\rm score_a}(M)$ given
in Section~\ref{sec:task}.

\begin{align}
{\rm score_r}(M)
& = \sum_{i=1}^{m} \sum_{j=i+1}^{m} {\rm sim_r}(a_i\!:\!a_j, M(a_i)\!:\!M(a_j)) \\
{\rm score_a}(M)
& = \sum_{i=1}^{m} {\rm sim_a}(a_i, M(a_i))
\end{align}

\noindent We can combine the scores by simply adding them or multiplying
them, but ${\rm score_r}(M)$ and ${\rm score_a}(M)$ may be quite different
in the scales and distributions of their values; therefore we first
normalize them to probabilities.

\begin{align}
{\rm prob_r}(M)
& = \frac{{\rm score_r}(M)}{\sum_{M_i \in P(A,B)} {\rm score_r}(M_i)} \\
{\rm prob_a}(M)
& = \frac{{\rm score_a}(M)}{\sum_{M_i \in P(A,B)} {\rm score_a}(M_i)}
\end{align}

\noindent For these probability estimates, we assume that ${\rm score_r}(M) \ge 0$
and ${\rm score_a}(M) \ge 0$. If necessary, a constant value may be added
to the scores, to ensure that they are not negative. Now we can combine
the scores by adding or multiplying the probabilities.

\begin{align}
M_{\rm r+a}
& = \operatornamewithlimits{arg\,max}_{M \in P(A,B)}
\big( {\rm prob_r}(M) + {\rm prob_a}(M) \big) \\
M_{\rm r \times a}
& = \operatornamewithlimits{arg\,max}_{M \in P(A,B)}
\big( {\rm prob_r}(M) \times {\rm prob_a}(M) \big)
\end{align}

Table~\ref{tab:hybrids} shows the accuracy when LRME is combined
with LIN+POS (the best attributional mapping algorithm in
Table~\ref{tab:attributional}, with an accuracy of 76.8\%)
or with HSO (the best attributional mapping algorithm that
does not use the manual POS tags, with an accuracy of 55.9\%).
We try both adding and multiplying probabilities. On its own,
LRME has an accuracy of 91.5\%. Combining LRME with LIN+POS
increases the accuracy to 94.0\%, but this improvement is not
statistically significant (paired t-test, 95\% confidence level).
Combining LRME with HSO results in a decrease in accuracy.
The decrease is not significant when the probabilities are
multiplied (85.4\%), but it is significant when the probabilities
are added (78.5\%).

\begin{table}[htbp]
\small
\centering
\begin{tabular}{lllr}
\hline
\multicolumn{2}{c}{\textbf{Components}} \\
\cline{1-2}
\textbf{Relational} & \textbf{Attributional} & \textbf{Combination} & \textbf{Accuracy} \\
\hline
LRME & LIN+POS & add probabilities      & 94.0 \\
LRME & LIN+POS & multiply probabilities & 94.0 \\
LRME & HSO     & add probabilities      & 78.5 \\
LRME & HSO     & multiply probabilities & 85.4 \\
\hline
\end{tabular}
\normalsize
\caption {The performance of four different hybrids of relational
and attributional mapping approaches.}
\label{tab:hybrids}
\end{table}

In summary, the experiments show no significant advantage to
combining LRME with attributional mapping. However, it is possible
that a larger sample of problems would show a significant advantage.
Also, the combination methods we explored (addition and multiplication
of probabilities) are elementary. A more sophisticated approach, such
as a weighted combination, may perform better.

\subsection{Coherent Relations}
\label{subsec:coherence}

We hypothesize that LRME benefits from a kind of coherence among the
relations. On the other hand, attributional mapping approaches do not
involve this kind of coherence.

Suppose we swap two of the terms in a mapping. Let $M$ be the original
mapping and let $M'$ be the new mapping, where $M'(a_1) = M(a_2)$,
$M'(a_2) = M(a_1)$, and $M'(a_i) = M(a_i)$ for $i > 2$. With attributional
similarity, the impact of this swap on the score of the mapping is limited.
Part of the score is not affected.

\begin{align}
{\rm score_a}(M) & = {\rm sim_a}(a_1, M(a_1)) + {\rm sim_a}(a_2, M(a_2))
+ \sum_{i=3}^{m} {\rm sim_a}(a_i, M(a_i)) \\
{\rm score_a}(M') & = {\rm sim_a}(a_1, M(a_2)) + {\rm sim_a}(a_2, M(a_1))
+ \sum_{i=3}^{m} {\rm sim_a}(a_i, M(a_i))
\end{align}

\noindent On the other hand, with relational similarity, the impact
of a swap is not limited in this way. A change to any part of the
mapping affects the whole score. There is a kind of global coherence
to relational similarity that is lacking in attributional similarity.

Testing the hypothesis that LRME benefits from coherence is somewhat
complicated, because we need to design the experiment so that the
coherence effect is isolated from any other effects. To do this, we
move some of the terms outside of the accuracy calculation.

Let $M_*: A \rightarrow B$ be one of our twenty mapping problems,
where $M_*$ is our intended mapping and
$m = \left | A \right | = \left | B \right |$. Let $A'$ be a randomly
selected subset of $A$ of size $m'$. Let $B'$ be $M_*(A')$, the subset
of $B$ to which $M_*$ maps $A'$.

\begin{align}
A' & \subset A \\
B' & \subset B \\
B' & = M_*(A') \\
m' & = \left | A' \right | = \left | B' \right | \\
m' & < m
\end{align}

\noindent There are two ways that we might use LRME to generate a mapping
$M': A' \rightarrow B'$ for this new reduced mapping problem,
{\em internal coherence} and {\em total coherence}.

\begin{myenumerate}

\item \textbf{Internal coherence:} We can select $M'$ based on
$\left \langle A', B' \right \rangle$ alone.

\begin{align}
A' & = \left \{ a_1, ..., a_{m'} \right \} \\
B' & = \left \{ b_1, ..., b_{m'} \right \} \\
M' & = \operatornamewithlimits{arg\,max}_{M \in P(A',B')}
\; \sum_{i=1}^{m'} \sum_{j=i+1}^{m'} {\rm sim_r}(a_i\!:\!a_j, M(a_i)\!:\!M(a_j))
\end{align}

\noindent In this case, $M'$ is chosen based only on the relations
that are internal to $\left \langle A', B' \right \rangle$.

\item \textbf{Total coherence:} We can select $M'$ based on
$\left \langle A, B \right \rangle$ and the knowledge that
$M'$ must satisfy the constraint that $M'(A') = B'$. (This knowledge
is also embedded in internal coherence.)

\begin{align}
A & = \left \{ a_1, ..., a_m \right \} \\
B & = \left \{ b_1, ..., b_m \right \} \\
P'(A,B) & = \left \{ M | \; M \in P(A,B) \; {\rm and} \; M(A') = B' \right \} \\
M' & = \operatornamewithlimits{arg\,max}_{M \in P'(A,B)}
\; \sum_{i=1}^{m} \sum_{j=i+1}^{m} {\rm sim_r}(a_i\!:\!a_j, M(a_i)\!:\!M(a_j))
\end{align}

\noindent In this case, $M'$ is chosen using both the relations
that are internal to $\left \langle A', B' \right \rangle$ and
other relations in $\left \langle A, B \right \rangle$ that are
external to  $\left \langle A', B' \right \rangle$.

\end{myenumerate}

Suppose that we calculate the accuracy of these two methods based
only on the sub\-problem $\left \langle A', B' \right \rangle$. At first
it might seem that there is no advantage to total coherence, because
it must explore a larger space of possible mappings than internal
coherence (since $\left | P'(A,B) \right |$ is larger
than $\left | P(A',B') \right |$),
but the additional terms that it explores are not involved in
calculating the accuracy. However, we hypothesize that total
coherence will have a higher accuracy than internal coherence,
because the additional external relations help to select the
correct mapping.

To test this hypothesis, we set $m'$ to 3 and we randomly generated
ten new reduced mapping problems for each of the twenty problems
(i.e., a total of 200 new problems of size 3). The average
accuracy of internal coherence was 93.3\%, whereas the average
accuracy of total coherence was 97.3\%. The difference is
statistically significant (paired t-test, 95\% confidence level).

On the other hand, the attributional mapping approaches cannot
benefit from total coherence, because there is no connection
between the attributes that are in $\left \langle A', B' \right \rangle$
and the attributes that are outside. We can decompose
${\rm score_a}(M)$ into two independent parts.

\begin{align}
A'' & = A \setminus A' \\
A   & = A' \cup A'' \\
P'(A,B) & = \left \{ M | \; M \in P(A,B) \; {\rm and} \; M(A') = B' \right \} \\
M'  & = \operatornamewithlimits{arg\,max}_{M \in P'(A,B)}
\; \sum_{a_i \in A} {\rm sim_a}(a_i, M(a_i)) \\
& = \operatornamewithlimits{arg\,max}_{M \in P'(A,B)}
\; \left ( \sum_{a_i \in A'} {\rm sim_a}(a_i, M(a_i)) +
\; \sum_{a_i \in A''} {\rm sim_a}(a_i, M(a_i)) \right )
\end{align}

\noindent These two parts can be optimized independently. Thus
the terms that are external to $\left \langle A', B' \right \rangle$
have no influence on the part of $M'$ that covers
$\left \langle A', B' \right \rangle$.

Relational mapping cannot be decomposed into independent parts in
this way, because the relations connect the parts. This gives relational
mapping approaches an inherent advantage over attributional mapping
approaches.

To confirm this analysis, we compared internal and total coherence
using LIN+POS on the same 200 new problems of size 3. The
average accuracy of internal coherence was 88.0\%, whereas the
average accuracy of total coherence was 87.0\%. The difference
is not statistically significant (paired t-test, 95\% confidence
level). (The only reason that there is any difference is that, when
two mappings have the same score, we break the ties randomly. This
causes random variation in the accuracy.)

The benefit from coherence suggests that we can make analogy
mapping problems easier for LRME by adding more terms. The
difficulty is that the new terms cannot be randomly chosen;
they must fit with the logic of the analogy and not overlap
with the existing terms.

Of course, this is not the only important difference between
the relational and attributional mapping approaches.
We believe that the most important difference is that relations
are more reliable and more general than
attributes, when using past experiences to make predictions
about the future \shortcite{hofstadter01,gentner03}.
Unfortunately, this hypothesis is more difficult
to evaluate experimentally than our hypothesis about coherence.

%
%

\section{Related Work}
\label{sec:related}

French \citeyear{french02} gives a good survey of computational
approaches to analogy-making, from the perspective of cognitive science
(where the emphasis is on how well computational systems model human
performance, rather than how well the systems perform). We will
sample a few systems from his survey and add a
few more that were not mentioned.

French \citeyear{french02} categorizes analogy-making systems as
{\em symbolic}, {\em connectionist}, or {\em symbolic-connectionist
hybrids}. G{\"a}rdenfors \citeyear{gardenfors04} proposes another
category of representational systems for AI and cognitive science, which he
calls {\em conceptual spaces}. These spatial or geometric
systems are common in information retrieval and machine learning
\shortcite{widdows04,rijsbergen04}. An influential example
is Latent Semantic Analysis \shortcite{landauer97}. The first
spatial approaches to analogy-making began to appear around the same
time as French's \citeyear{french02} survey. LRME takes a spatial
approach to analogy-making.

\subsection{Symbolic Approaches}

Computational approaches to analogy-making date back to {\sc Analogy}
\shortcite{evans64} and Argus \shortcite{reitman65}. Both of these
systems were designed to solve proportional analogies (analogies
in which $\left | A \right | = \left | B \right | = 2$; see
Section~\ref{sec:lra}). {\sc Analogy} could solve proportional
analogies with simple geometric figures and Argus could solve
simple word analogies. These systems used hand-coded rules
and were only able to solve the limited range of problems that
their designers had anticipated and coded in the rules.

French \citeyear{french02} cites Structure Mapping Theory (SMT)
\shortcite{gentner83} and the Structure Mapping Engine (SME)
\shortcite{falkenhainer89} as the prime examples of symbolic
approaches:

\begin{quote}

SMT is unquestionably the most influential work to date on the
modeling of analogy-making and has been applied in a wide range
of contexts ranging from child development to folk physics.
SMT explicitly shifts the emphasis in analogy-making to
the structural similarity between the source and target domains.
Two major principles underlie SMT:

\begin{myitemize}

\item the relation-matching principle: good analogies are determined by
mappings of relations and not attributes (originally only identical
predicates were mapped) and

\item the systematicity principle: mappings of coherent systems of
relations are preferred over mappings of individual relations.

\end{myitemize}

This structural approach was intended to produce a domain-independent
mapping process.

\end{quote}

\noindent LRME follows both of these principles. LRME uses only
relational similarity; no attributional similarity is involved
(see Section~\ref{subsec:algorithm}). Coherent systems of relations
are preferred over mappings of individual relations (see
Section~\ref{subsec:coherence}). However, the spatial (statistical,
corpus-based) approach of LRME is quite different from the symbolic
(logical, hand-coded) approach of SME.

Martin \citeyear{martin92} uses a symbolic approach to handle
conventional metaphors. Gentner, Bowdle, Wolff, and Boronat
\citeyear{gentner01} argue that novel metaphors are processed as
analogies, but conventional metaphors are recalled from memory without
special processing. However, the line between conventional and novel
metaphor can be unclear.

Dolan \citeyear{dolan95} describes an algorithm that can
extract conventional metaphors from a dictionary. A semantic parser is
used to extract semantic relations from the Longman Dictionary of
Contemporary English (LDOCE). A symbolic algorithm finds
metaphorical relations between words, using the extracted
relations.

Veale \citeyear{veale03,veale04} has developed a symbolic approach
to analogy-making, using WordNet as a lexical resource. Using
a spreading activation algorithm, he achieved a score of 43.0\%
on a set of 374 multiple-choice lexical proportional analogy
questions from the SAT college entrance test \shortcite{veale04}.

Lepage \citeyear{lepage98} has demonstrated that a symbolic
approach to proportional analogies can be used for morphology
processing. Lepage and Denoual \citeyear{lepage05} apply
a similar approach to machine translation.

\subsection{Connectionist Approaches}

Connectionist approaches to analogy-making include ACME \shortcite{holyoak89}
and LISA \shortcite{hummel97}. Like symbolic approaches, these systems
use hand-coded knowledge representations, but the search for mappings
takes a connectionist approach, in which there are nodes with weights
that are incrementally updated over time, until the system reaches
a stable state.

\subsection{Symbolic-Connectionist Hybrid Approaches}

The third family examined by French \citeyear{french02} is
hybrid approaches, containing elements of both the
symbolic and connectionist approaches. Examples include
Copycat \shortcite{mitchell93} and Tabletop
\shortcite{french95}. Much of the work in the
Fluid Analogies Research Group (FARG) concerns
symbolic-connectionist hybrids \shortcite{hofstadter95}.

\subsection{Spatial Approaches}

Marx, Dagan, Buhmann, and Shamir \citeyear{marx02} present the
{\em coupled clustering} algorithm, which uses a feature vector
representation to find analogies in collections of text.
For example, given documents on Buddhism and Christianity,
it finds related terms, such as \{{\em school, Mahayana, Zen}\}
for Buddhism and \{{\em tradition, Catholic, Protestant}\} for
Christianity.

Mason \citeyear{mason04} describes the CorMet system
for extracting conventional metaphors from text.
CorMet is based on clustering feature vectors that
represent the selectional preferences of verbs. Given
keywords for the source domain {\em laboratory} and
the target domain {\em finance}, it is able to discover
mappings such as {\em liquid} $\rightarrow$ {\em income}
and {\em container} $\rightarrow$ {\em institution}.

Turney, Littman, Bigham, and Shnayder \citeyear{turney03} present
a system for solving lexical proportional analogy
questions from the SAT college entrance test, which
combines thirteen different modules. Twelve of the
modules use either attributional similarity or
a symbolic approach to relational similarity, but one
module uses a spatial (feature vector) approach to
measuring relational similarity. This module worked
much better than any of the other modules; therefore,
it was studied in more detail by Turney and Littman
\citeyear{turney05a}. The relation between a pair of
words is represented by a vector, in which the elements are
pattern frequencies. This is similar to LRME, but one important
difference is that Turney and Littman \citeyear{turney05a} used
a fixed, hand-coded set of 128 patterns, whereas LRME
automatically generates a variable number of patterns from the
given corpus (33,240 patterns in our experiments here).

Turney \citeyear{turney05b} introduced Latent Relational
Analysis (LRA), which was examined more thoroughly
by Turney \citeyear{turney06}. LRA achieves human-level
performance on a set of 374 multiple-choice proportional
analogy questions from the SAT college entrance exam.
LRME uses a simplified form of LRA. A similar simplification
of LRA is used by Turney \citeyear{turney08}, in a system for
processing analogies, synonyms, antonyms, and associations.
The contribution of LRME is to go beyond proportional
analogies, to larger systems of analogical mappings.

\subsection{General Theories of Analogy and Metaphor}

Many theories of analogy-making and metaphor either do not involve
computation or they suggest general principles and concepts that are
not specific to any particular computational approach.
The design of LRME has been influenced by several theories
of this type \shortcite{gentner83,hofstadter95,holyoak95,hofstadter01,gentner03}.

Lakoff and Johnson \citeyear{lakoff80} provide extensive evidence
that metaphor is ubiquitous in language and thought. We believe that
a system for analogy-making should be able to handle metaphorical
language, which is why ten of our analogy problems are derived
from Lakoff and Johnson \citeyear{lakoff80}. We agree with
their claim that a metaphor does not merely involve a
superficial relation between a couple of words; rather, it
involves a systematic set of mappings between two domains.
Thus our analogy problems involve larger sets of words, beyond
proportional analogies.

Holyoak and Thagard \citeyear{holyoak95} argue that analogy-making
is central in our daily thought, and especially in finding
creative solutions to new problems. Our ten scientific analogies
were derived from their examples of analogy-making in
scientific creativity.

%
%

\section{Limitations and Future Work}
\label{sec:future}

In Section~\ref{sec:lra}, we mentioned ten applications for LRA,
and in Section~\ref{sec:apps} we claimed that the results of
the experiments in Section~\ref{subsec:coherence} suggest that
LRME may perform better than LRA on all ten of these applications,
due to its ability to handle bijective analogies when $m > 2$.
Our focus in future work will be testing this hypothesis. In
particular, the task of semantic role labeling, discussed in
Section~\ref{sec:intro}, seems to be a good candidate application
for LRME.

The input to LRME is simpler than the input to SME (compare
Figures \ref{fig:solar} and \ref{fig:atom} in Section~\ref{sec:intro}
with Table~\ref{tab:input}), but there is still some human effort involved
in creating the input. LRME is not immune to the criticism of Chalmers,
French, and Hofstadter \citeyear{chalmers92}, that the human who generates
the input is doing more work than the computer that makes the mappings,
although it is not a trivial matter to find the right
mapping out of 5,040 (7!) choices.

In future work, we would like to relax the requirement that
$\left \langle A, B \right \rangle$ must be a bijection (see
Section~\ref{sec:task}), by adding irrelevant words (distractors)
and synonyms. The mapping algorithm will be forced to decide
what terms to include in the mapping and what terms to leave out.

We would also like to develop an algorithm that can take a proportional
analogy ($m = 2$) as input (e.g., sun:planet::nucleus:electron) and automatically
expand it to a larger analogy ($m > 2$, e.g., Table~\ref{tab:output}).
That is, it would automatically search the corpus for new
terms to add to the analogy.

The next step would be to give the computer only the topic
of the source domain (e.g., solar system) and the topic of
the target domain (e.g., atomic structure), and let it work
out the rest on its own. This might be possible by combining
ideas from LRME with ideas from coupled clustering
\shortcite{marx02} and CorMet \shortcite{mason04}.

It seems that analogy-making is triggered in people when we
encounter a problem \shortcite{holyoak95}. The problem defines
the target for us, and we immediately start searching for
a source. Analogical mapping enables us to transfer our knowledge
of the source to the target, hopefully leading to a solution to
the problem. This suggests that the input to the ideal
analogical mapping algorithm would be simply a statement
that there is a problem (e.g., What is the structure of the
atom?). Ultimately, the computer might find the problems on its
own as well. The only input would be a large corpus.

The algorithms we have considered here all perform exhaustive
search of the set of possible mappings $P(A,B)$. This is
acceptable when the sets are small, as they are here, but
it will be problematic for larger problems. In future work,
it will be necessary to use heuristic search algorithms
instead of exhaustive search.

It takes almost 18 hours for LRME to process the twenty
mapping problems (Section~\ref{sec:lrme}). With better
hardware and some changes to the software, this time
could be significantly reduced. For even greater speed, the
algorithm could run continuously, building a large
database of vector representations of term pairs,
so that it is ready to create mappings as soon
as a user requests them. This is similar to the vision
of Banko and Etzioni \citeyear{banko07}.

LRME, like LRA and LSA \shortcite{landauer97}, uses a truncated singular
value decomposition (SVD) to smooth the matrix. Many other algorithms
have been proposed for smoothing matrices. In our past work
with LRA \shortcite{turney06}, we experimented with Nonnegative
Matrix Factorization (NMF) \shortcite{lee99}, Probabilistic
Latent Semantic Analysis (PLSA) \shortcite{hofmann99},
Iterative Scaling (IS) \shortcite{ando00}, and Kernel Principal
Components Analysis (KPCA) \cite{scholkopf97}. We had some
interesting results with small matrices (around 1000 $\times$ 2000),
but none of the algorithms seemed substantially better than truncated
SVD, and none of them scaled up to the matrix sizes that we have
here (1,662 $\times$ 33,240). However, we believe that SVD is not
unique, and future work is likely to discover a
smoothing algorithm that is more efficient and effective than SVD.
The results in Section~\ref{subsec:experiments} do not show
a significant benefit from SVD. Table~\ref{tab:lrme-variations}
hints that PPMIC \shortcite{bullinaria07} is more important than SVD.

LRME extracts knowledge from many fragments of text. In
Section~\ref{subsec:algorithm}, we noted that we found an average
of 1,180 phrases per pair. The information from these 1,180 phrases
is combined in a vector, to represent the semantic relation for
a pair. This is quite different from relation extraction in (for
example) the Automatic Content Extraction (ACE)
Evaluation.\footnote{ACE is an annual event that began in 1999.
Relation Detection and Characterization (RDC) was introduced to ACE
in 2001. For more information, see http://www.nist.gov/speech/tests/ace/.}
The task in ACE is to identify and label a semantic relation in
a single sentence. Semantic role labeling also involves labeling
a single sentence \shortcite{gildea02}.

The contrast between LRME and ACE is analogous to the distinction
in cognitive psychology between semantic and episodic
memory. Episodic memory is memory of a specific event in one's personal past,
whereas semantic memory is memory of basic facts and concepts,
unrelated to any specific event in the past. LRME extracts relational
information that is independent of any specific sentence, like semantic
memory. ACE is concerned with extracting the relation in a specific
sentence, like episodic memory. In cognition, episodic memory and
semantic memory work together synergistically. When we experience
an event, we use our semantic memory to interpret the event and
form a new episodic memory, but semantic memory is itself constructed
from our past experiences, our accumulated episodic memories. This
suggests that there should be a synergy from combining LRME-like
semantic information extraction algorithms with ACE-like
episodic information extraction algorithms.

%
%

\section{Conclusion}
\label{sec:conclusion}

Analogy is the core of cognition. We understand the present
by analogy to the past. We predict the future by analogy
to the past and the present. We solve problems by searching
for analogous situations \shortcite{holyoak95}. Our daily language
is saturated with metaphor \shortcite{lakoff80}, and metaphor
is based on analogy \shortcite{gentner01}. To understand
human language, to solve human problems, to work with humans,
computers must be able to make analogical mappings.

Our best theory of analogy-making is Structure Mapping Theory
\shortcite{gentner83}, but the Structure Mapping Engine
\shortcite{falkenhainer89} puts too much of the burden
of analogy-making on its human users \shortcite{chalmers92}.
LRME is an attempt to shift some of that burden onto the
computer, while remaining consistent with the general principles
of SMT.

We have shown that LRME is able to solve bijective analogical
mapping problems with human-level performance. Attributional
mapping algorithms (at least, those we have tried so far) are
not able to reach this level. This supports SMT, which claims
that relations are more important than attributes when making
analogical mappings.

There is still much research to be done. LRME takes some of
the load off the human user, but formulating the input to LRME
is not easy. This paper is an incremental step
towards a future in which computers can make surprising and
useful analogies with minimal human assistance.

%
%

\acks{Thanks to my colleagues at the Institute for Information
Technology for participating in the experiment in
Section~\ref{sec:problems}. Thanks to Charles Clarke and
Egidio Terra for their corpus. Thanks to Stefan B{\"u}ttcher
for making Wumpus available and giving me advice on its use.
Thanks to Doug Rohde for making SVDLIBC available. Thanks to
the WordNet team at Princeton University for WordNet, Ted Pedersen
for the WordNet::Similarity Perl package, and Jason Rennie for
the WordNet::QueryData Perl package. Thanks to the LSA team
at the University of Colorado at Boulder for the use of their
online demonstration of LSA. Thanks to Deniz Yuret, Andr{\'e}
Vellino, Dedre Gentner, Vivi Nastase, Yves Lepage, Diarmuid {\'O}
S{\'e}aghdha, Roxana Girju, Chris Drummond, Howard Johnson, Stan
Szpakowicz, and the anonymous reviewers of {\em JAIR} for their helpful
comments and suggestions.}

%
%

\appendix
\section*{Appendix A. Details of the Mapping Problems}

In this appendix, we provide detailed information about
the twenty mapping problems.
Figure \ref{fig:instructions} shows the instructions that
were given to the participants in the experiment
in Section~\ref{sec:problems}. These instructions were
displayed in their web browsers.
Tables \ref{tab:a1-a5}, \ref{tab:a6-a10}, \ref{tab:m1-m5}, and
\ref{tab:m6-m10} show the twenty mapping problems. The first column
gives the problem number (e.g., A1) and a mnemonic that summarizes
the mapping (e.g., solar system $\rightarrow$ atom). The second column
gives the source terms and the third column gives the target terms.

\begin{table}[htbp]
\centering
\begin{tabular}{|p{\textwidth}|}
\hline
\textbf{Systematic Analogies and Metaphors} \\
\\
\textbf{Instructions} \\
\small
\\
You will be presented with twenty analogical mapping problems, ten
based on scientific analogies and ten based on common metaphors.
A typical problem will look like this: \\
\\
\hspace{1in} \makebox[1in][l]{horse} $\rightarrow$
\framebox[1in][l]{? \hfill $\nabla$} \\
\hspace{1in} \makebox[1in][l]{legs}  $\rightarrow$
\framebox[1in][l]{? \hfill $\nabla$} \\
\hspace{1in} \makebox[1in][l]{hay}   $\rightarrow$
\framebox[1in][l]{? \hfill $\nabla$} \\
\hspace{1in} \makebox[1in][l]{brain} $\rightarrow$
\framebox[1in][l]{? \hfill $\nabla$} \\
\hspace{1in} \makebox[1in][l]{dung}  $\rightarrow$
\framebox[1in][l]{? \hfill $\nabla$} \\
\\
You may click on the drop-down menus above, to see what
options are available. \\
\\
Your task is to construct an analogical mapping; that
is, a one-to-one mapping between the items on the left
and the items on the right. For example: \\
\\
\hspace{1in} \makebox[1in][l]{horse} $\rightarrow$
\framebox[1in][l]{car \hfill $\nabla$} \\
\hspace{1in} \makebox[1in][l]{legs}  $\rightarrow$
\framebox[1in][l]{wheels \hfill $\nabla$} \\
\hspace{1in} \makebox[1in][l]{hay}   $\rightarrow$
\framebox[1in][l]{gasoline \hfill $\nabla$} \\
\hspace{1in} \makebox[1in][l]{brain} $\rightarrow$
\framebox[1in][l]{driver \hfill $\nabla$} \\
\hspace{1in} \makebox[1in][l]{dung}  $\rightarrow$
\framebox[1in][l]{exhaust \hfill $\nabla$} \\
\\
This mapping expresses an analogy between a horse and
a car. The horse's legs are like the car's wheels. The
horse eats hay and the car consumes gasoline. The horse's
brain controls the movement of the horse like the car's driver
controls the movement of the car. The horse generates dung
as a waste product like the car generates exhaust as a waste
product. \\
\\
You should have no duplicate items in your answers on the
right-hand side. If there are any duplicates or missing
items (question marks), you will get an error message when
you submit your answer. \\
\\
You are welcome to use a dictionary as you work on
the problems, if you would find it helpful. \\
\\
If you find the above instructions unclear,
then please do not continue
with this exercise. Your answers to the twenty problems will
be used as a standard for evaluating the output of a computer
algorithm; therefore, you should only proceed if you are
confident that you understand this task. \\
\hline
\end{tabular}
\normalsize
\figcaption {The instructions for the participants
in the experiment in Section~\ref{sec:problems}.}
\label{fig:instructions}
\end{table}

\begin{table}[htbp]
\small
\centering
\begin{tabular}{lllrl}
\hline
\textbf{Mapping} & \textbf{Source} & \textbf{$\rightarrow$ Target}
& \textbf{Agreement} & \textbf{POS} \\
\hline
& solar system         & $\rightarrow$ atom                 &  86.4 & NN  \\
A1
& sun                  & $\rightarrow$ nucleus              & 100.0 & NN  \\
& planet               & $\rightarrow$ electron             &  95.5 & NN  \\
solar system
& mass                 & $\rightarrow$ charge               &  86.4 & NN  \\
$\rightarrow$ atom
& attracts             & $\rightarrow$ attracts             &  90.9 & VBZ \\
& revolves             & $\rightarrow$ revolves             &  95.5 & VBZ \\
& gravity              & $\rightarrow$ electromagnetism     &  81.8 & NN  \\
\cline{2-5}
& Average agreement: & &  90.9 & \\
\hline
& water                & $\rightarrow$ heat                 &  86.4 & NN  \\
A2
& flows                & $\rightarrow$ transfers            &  95.5 & VBZ \\
& pressure             & $\rightarrow$ temperature          &  86.4 & NN  \\
water flow
& water tower          & $\rightarrow$ burner               &  72.7 & NN  \\
$\rightarrow$ heat transfer
& bucket               & $\rightarrow$ kettle               &  72.7 & NN  \\
& filling              & $\rightarrow$ heating              &  95.5 & VBG \\
& emptying             & $\rightarrow$ cooling              &  95.5 & VBG \\
& hydrodynamics        & $\rightarrow$ thermodynamics       &  90.9 & NN  \\
\cline{2-5}
& Average agreement: & &  86.9 & \\
\hline
& waves                & $\rightarrow$ sounds               &  86.4 & NNS \\
A3
& shore                & $\rightarrow$ wall                 &  77.3 & NN  \\
& reflects             & $\rightarrow$ echoes               &  95.5 & VBZ \\
waves
& water                & $\rightarrow$ air                  &  95.5 & NN  \\
$\rightarrow$ sounds
& breakwater           & $\rightarrow$ insulation           &  81.8 & NN  \\
& rough                & $\rightarrow$ loud                 &  63.6 & JJ  \\
& calm                 & $\rightarrow$ quiet                & 100.0 & JJ  \\
& crashing             & $\rightarrow$ vibrating            &  54.5 & VBG \\
\cline{2-5}
& Average agreement: & &  81.8 & \\
\hline
& combustion           & $\rightarrow$ respiration          &  72.7 & NN  \\
A4
& fire                 & $\rightarrow$ animal               &  95.5 & NN  \\
& fuel                 & $\rightarrow$ food                 &  90.9 & NN  \\
combustion
& burning              & $\rightarrow$ breathing            &  72.7 & VBG \\
$\rightarrow$ respiration
& hot                  & $\rightarrow$ living               &  59.1 & JJ  \\
& intense              & $\rightarrow$ vigorous             &  77.3 & JJ  \\
& oxygen               & $\rightarrow$ oxygen               &  77.3 & NN  \\
& carbon dioxide       & $\rightarrow$ carbon dioxide       &  86.4 & NN  \\
\cline{2-5}
& Average agreement: & &  79.0 & \\
\hline
& sound                & $\rightarrow$ light                &  86.4 & NN  \\
A5
& low                  & $\rightarrow$ red                  &  50.0 & JJ  \\
& high                 & $\rightarrow$ violet               &  54.5 & JJ  \\
sound
& echoes               & $\rightarrow$ reflects             & 100.0 & VBZ \\
$\rightarrow$ light
& loud                 & $\rightarrow$ bright               &  90.9 & JJ  \\
& quiet                & $\rightarrow$ dim                  &  77.3 & JJ  \\
& horn                 & $\rightarrow$ lens                 &  95.5 & NN  \\
\cline{2-5}
& Average agreement: & &  79.2 & \\
\hline
\end{tabular}
\normalsize
\caption {Science analogy problems A1 to A5, derived from Chapter~8 of
Holyoak and Thagard \citeyear{holyoak95}.}
\label{tab:a1-a5}
\end{table}

\begin{table}[htbp]
\small
\centering
\begin{tabular}{lllrl}
\hline
\textbf{Mapping} & \textbf{Source} & \textbf{$\rightarrow$ Target}
& \textbf{Agreement} & \textbf{POS} \\
\hline
& projectile           & $\rightarrow$ planet               & 100.0 & NN  \\
A6
& trajectory           & $\rightarrow$ orbit                & 100.0 & NN  \\
& earth                & $\rightarrow$ sun                  & 100.0 & NN  \\
projectile
& parabolic            & $\rightarrow$ elliptical           & 100.0 & JJ  \\
$\rightarrow$ planet
& air                  & $\rightarrow$ space                & 100.0 & NN  \\
& gravity              & $\rightarrow$ gravity              &  90.9 & NN  \\
& attracts             & $\rightarrow$ attracts             &  90.9 & VBZ \\
\cline{2-5}
& Average agreement: & &  97.4 & \\
\hline
& breeds               & $\rightarrow$ species              & 100.0 & NNS \\
A7
& selection            & $\rightarrow$ competition          &  59.1 & NN  \\
& conformance          & $\rightarrow$ adaptation           &  59.1 & NN  \\
artificial selection
& artificial           & $\rightarrow$ natural              &  77.3 & JJ  \\
$\rightarrow$ natural selection
& popularity           & $\rightarrow$ fitness              &  54.5 & NN  \\
& breeding             & $\rightarrow$ mating               &  95.5 & VBG \\
& domesticated         & $\rightarrow$ wild                 &  77.3 & JJ  \\
\cline{2-5}
& Average agreement: & &  74.7 & \\
\hline
& balls                & $\rightarrow$ molecules            &  90.9 & NNS \\
A8
& billiards            & $\rightarrow$ gas                  &  72.7 & NN  \\
& speed                & $\rightarrow$ temperature          &  81.8 & NN  \\
billiard balls
& table                & $\rightarrow$ container            &  95.5 & NN  \\
$\rightarrow$ gas molecules
& bouncing             & $\rightarrow$ pressing             &  77.3 & VBG \\
& moving               & $\rightarrow$ moving               &  86.4 & VBG \\
& slow                 & $\rightarrow$ cold                 & 100.0 & JJ  \\
& fast                 & $\rightarrow$ hot                  & 100.0 & JJ  \\
\cline{2-5}
& Average agreement: & &  88.1 & \\
\hline
& computer             & $\rightarrow$ mind                 &  90.9 & NN  \\
A9
& processing           & $\rightarrow$ thinking             &  95.5 & VBG \\
& erasing              & $\rightarrow$ forgetting           & 100.0 & VBG \\
computer
& write                & $\rightarrow$ memorize             &  72.7 & VB  \\
$\rightarrow$ mind
& read                 & $\rightarrow$ remember             &  54.5 & VB  \\
& memory               & $\rightarrow$ memory               &  81.8 & NN  \\
& outputs              & $\rightarrow$ muscles              &  72.7 & NNS \\
& inputs               & $\rightarrow$ senses               &  90.9 & NNS \\
& bug                  & $\rightarrow$ mistake              & 100.0 & NN  \\
\cline{2-5}
& Average agreement: & &  84.3 & \\
\hline
& slot machines        & $\rightarrow$ bacteria             &  68.2 & NNS \\
A10
& reels                & $\rightarrow$ genes                &  72.7 & NNS \\
& spinning             & $\rightarrow$ mutating             &  86.4 & VBG \\
slot machine
& winning              & $\rightarrow$ reproducing          &  90.9 & VBG \\
$\rightarrow$ bacterial mutation
& losing               & $\rightarrow$ dying                & 100.0 & VBG \\
\cline{2-5}
& Average agreement: & &  83.6 & \\
\hline
\end{tabular}
\normalsize
\caption {Science analogy problems A6 to A10, derived from Chapter~8 of
Holyoak and Thagard \citeyear{holyoak95}.}
\label{tab:a6-a10}
\end{table}

\begin{table}[htbp]
\small
\centering
\begin{tabular}{lllrl}
\hline
\textbf{Mapping} & \textbf{Source} & \textbf{$\rightarrow$ Target}
& \textbf{Agreement} & \textbf{POS} \\
\hline
& war                  & $\rightarrow$ argument             &  90.9 & NN  \\
M1
& soldier              & $\rightarrow$ debater              & 100.0 & NN  \\
& destroy              & $\rightarrow$ refute               &  90.9 & VB  \\
war
& fighting             & $\rightarrow$ arguing              &  95.5 & VBG \\
$\rightarrow$ argument
& defeat               & $\rightarrow$ acceptance           &  90.9 & NN  \\
& attacks              & $\rightarrow$ criticizes           &  95.5 & VBZ \\
& weapon               & $\rightarrow$ logic                &  90.9 & NN  \\
\cline{2-5}
& Average agreement: & &  93.5 & \\
\hline
& buyer                & $\rightarrow$ believer             & 100.0 & NN  \\
M2
& merchandise          & $\rightarrow$ belief               &  90.9 & NN  \\
& buying               & $\rightarrow$ accepting            &  95.5 & VBG \\
buying an item
& selling              & $\rightarrow$ advocating           & 100.0 & VBG \\
$\rightarrow$ accepting a belief
& returning            & $\rightarrow$ rejecting            &  95.5 & VBG \\
& valuable             & $\rightarrow$ true                 &  95.5 & JJ  \\
& worthless            & $\rightarrow$ false                &  95.5 & JJ  \\
\cline{2-5}
& Average agreement: & &  96.1 & \\
\hline
& foundations          & $\rightarrow$ reasons              &  72.7 & NNS \\
M3
& buildings            & $\rightarrow$ theories             &  77.3 & NNS \\
& supporting           & $\rightarrow$ confirming           &  95.5 & VBG \\
grounds for a building
& solid                & $\rightarrow$ rational             &  90.9 & JJ  \\
$\rightarrow$ reasons for a theory
& weak                 & $\rightarrow$ dubious              &  95.5 & JJ  \\
& crack                & $\rightarrow$ flaw                 &  95.5 & NN  \\
\cline{2-5}
& Average agreement: & &  87.9 & \\
\hline
& obstructions         & $\rightarrow$ difficulties         & 100.0 & NNS \\
M4
& destination          & $\rightarrow$ goal                 & 100.0 & NN  \\
& route                & $\rightarrow$ plan                 & 100.0 & NN  \\
impediments to travel
& traveller            & $\rightarrow$ person               & 100.0 & NN  \\
$\rightarrow$ difficulties
& travelling           & $\rightarrow$ problem solving      & 100.0 & VBG \\
& companion            & $\rightarrow$ partner              & 100.0 & NN  \\
& arriving             & $\rightarrow$ succeeding           & 100.0 & VBG \\
\cline{2-5}
& Average agreement: & & 100.0 & \\
\hline
& money                & $\rightarrow$ time                 &  95.5 & NN  \\
M5
& allocate             & $\rightarrow$ invest               &  86.4 & VB  \\
& budget               & $\rightarrow$ schedule             &  86.4 & NN  \\
money
& effective            & $\rightarrow$ efficient            &  86.4 & JJ  \\
$\rightarrow$ time
& cheap                & $\rightarrow$ quick                &  50.0 & JJ  \\
& expensive            & $\rightarrow$ slow                 &  59.1 & JJ  \\
\cline{2-5}
& Average agreement: & &  77.3 & \\
\hline
\end{tabular}
\normalsize
\caption {Common metaphor problems M1 to M5, derived from Lakoff and
Johnson \citeyear{lakoff80}.}
\label{tab:m1-m5}
\end{table}

\begin{table}[htbp]
\small
\centering
\begin{tabular}{lllrl}
\hline
\textbf{Mapping} & \textbf{Source} & \textbf{$\rightarrow$ Target}
& \textbf{Agreement} & \textbf{POS} \\
\hline
& seeds                & $\rightarrow$ ideas                &  90.9 & NNS \\
M6
& planted              & $\rightarrow$ inspired             &  95.5 & VBD \\
& fruitful             & $\rightarrow$ productive           &  81.8 & JJ  \\
seeds
& fruit                & $\rightarrow$ product              &  95.5 & NN  \\
$\rightarrow$ ideas
& grow                 & $\rightarrow$ develop              &  81.8 & VB  \\
& wither               & $\rightarrow$ fail                 & 100.0 & VB  \\
& blossom              & $\rightarrow$ succeed              &  77.3 & VB  \\
\cline{2-5}
& Average agreement: & &  89.0 & \\
\hline
& machine              & $\rightarrow$ mind                 &  95.5 & NN  \\
M7
& working              & $\rightarrow$ thinking             & 100.0 & VBG \\
& turned on            & $\rightarrow$ awake                & 100.0 & JJ  \\
machine
& turned off           & $\rightarrow$ asleep               & 100.0 & JJ  \\
$\rightarrow$ mind
& broken               & $\rightarrow$ confused             & 100.0 & JJ  \\
& power                & $\rightarrow$ intelligence         &  95.5 & NN  \\
& repair               & $\rightarrow$ therapy              & 100.0 & NN  \\
\cline{2-5}
& Average agreement: & &  98.7 & \\
\hline
& object               & $\rightarrow$ idea                 &  90.9 & NN  \\
M8
& hold                 & $\rightarrow$ understand           &  81.8 & VB  \\
& weigh                & $\rightarrow$ analyze              &  81.8 & VB  \\
object
& heavy                & $\rightarrow$ important            &  95.5 & JJ  \\
$\rightarrow$ idea
& light                & $\rightarrow$ trivial              &  95.5 & JJ  \\
\cline{2-5}
& Average agreement: & &  89.1 & \\
\hline
& follow               & $\rightarrow$ understand           & 100.0 & VB  \\
M9
& leader               & $\rightarrow$ speaker              & 100.0 & NN  \\
& path                 & $\rightarrow$ argument             & 100.0 & NN  \\
following
& follower             & $\rightarrow$ listener             & 100.0 & NN  \\
$\rightarrow$ understanding
& lost                 & $\rightarrow$ misunderstood        &  86.4 & JJ  \\
& wanders              & $\rightarrow$ digresses            &  90.9 & VBZ \\
& twisted              & $\rightarrow$ complicated          &  95.5 & JJ  \\
& straight             & $\rightarrow$ simple               & 100.0 & JJ  \\
\cline{2-5}
& Average agreement: & &  96.6 & \\
\hline
& seeing               & $\rightarrow$ understanding        &  68.2 & VBG \\
M10
& light                & $\rightarrow$ knowledge            &  77.3 & NN  \\
& illuminating         & $\rightarrow$ explaining           &  86.4 & VBG \\
seeing
& darkness             & $\rightarrow$ confusion            &  86.4 & NN  \\
$\rightarrow$ understanding
& view                 & $\rightarrow$ interpretation       &  68.2 & NN  \\
& hidden               & $\rightarrow$ secret               &  86.4 & JJ  \\
\cline{2-5}
& Average agreement: & &  78.8 & \\
\hline
\end{tabular}
\normalsize
\caption {Common metaphor problems M6 to M10, derived from Lakoff and
Johnson \citeyear{lakoff80}.}
\label{tab:m6-m10}
\end{table}

The mappings shown in these tables are our intended
mappings. The fourth column shows the percentage of participants
who agreed with our intended mappings. For example, in
problem A1, 81.8\% of the participants (18 out of 22) mapped
gravity to electromagnetism. The final column gives the
part-of-speech (POS) tags for the source and target terms. We used
the Penn Treebank tags \shortcite{santorini90}. We assigned
these tags manually. Our intended mappings and our tags were chosen
so that mapped terms have the same tags. For example, in A1, {\em sun}
maps to {\em nucleus}, and both {\em sun} and {\em nucleus} are tagged NN.
The POS tags are used in the experiments in Section~\ref{sec:attributes}.
The POS tags are not used by LRME and they were not shown to the
participants in the experiment in Section~\ref{sec:problems}.

\vskip 0.2in

\bibliography{NRC-50738}
\bibliographystyle{theapa}

\end{document}